\documentclass{article}

 \usepackage[main, final]{neurips_2026}
\usepackage[utf8]{inputenc} % allow utf-8 input
\usepackage[T1]{fontenc}    % use 8-bit T1 fonts
\usepackage{hyperref}       % hyperlinks
\usepackage{url}            % simple URL typesetting
\usepackage{booktabs}       % professional-quality tables
\usepackage{amsfonts}       % blackboard math symbols
\usepackage{nicefrac}       % compact symbols for 1/2, etc.
\usepackage{microtype}      % microtypography
\usepackage{xcolor}         % colors

\usepackage{microtype}
\usepackage{graphicx}
\usepackage{subcaption}

\usepackage{amsmath}
\usepackage{amssymb}
\usepackage{mathtools}
\usepackage{amsthm}
\usepackage{comment}

\usepackage[capitalize,noabbrev]{cleveref}
\usepackage{tabularx,ragged2e}
\usepackage{titletoc}

\newcommand\DoToC{%
  \startcontents
  \printcontents{}{1}{\textbf{Table of Contents}\vskip3pt\hrule\vskip5pt}
  \vskip3pt\hrule\vskip5pt
}

\theoremstyle{plain}
\newtheorem{theorem}{Theorem}[section]
\newtheorem{proposition}[theorem]{Proposition}
\newtheorem{lemma}[theorem]{Lemma}
\newtheorem{corollary}[theorem]{Corollary}
\newtheorem{assumption}[theorem]{Assumption}
\theoremstyle{definition}
\newtheorem{definition}[theorem]{Definition}
\newtheorem{conjecture}[theorem]{Conjecture}
\newtheorem{remark}[theorem]{Remark}

\def\eps{\varepsilon}

\def\bega{\begin{aligned}}
\def\enda{\end{aligned}}
\def\R{\mathbb{R}^2}

\def\lw{\left}
\def\rw{\right}

\def\R{\mathbb{R}}

\def\la{\langle}
\def\ra{\rangle}

\def\bcase{\begin{cases}}
\def\ecase{\end{cases}}
\def\al{\alpha}
\def\bmx{\begin{bmatrix}}
\def\emx{\end{bmatrix}}

\def\bitem{\begin{itemize}}
\def\eitem{\end{itemize}}

\newcommand{\Be}{\begin{equation}}
	\newcommand{\Ee}{\end{equation}}

\renewcommand{\P}{\mathbf{P}}
\renewcommand{\S}{\mathbb{S}}
\newcommand{\be}{\begin{equation}}
\newcommand{\bm}{\begin{multline}}
\newcommand{\ee}{\end{equation}}

\newcommand{\tp}{^{\top}}

\newcommand{\Bes}{\begin{eqnarray*}}
\newcommand{\Ees}{\end{eqnarray*}}

\newcommand{\Bs}{\begin{split}}

\def\beq{\begin{equation}}
\def\eeq{\end{equation}}
\def\bb1{{1\!\!1}}

\def\R{\mathbb{R}}
\def\pt{\partial}

\def\diag{\mbox{diag}}

\title{Two-Clustering Regime of Token Dynamics\\ in Causal Attention}

\author{%
  Trinh T. Nguyen\\
  School of Mathematical Sciences\\
  Shanghai Jiao Tong University, China\\
  \texttt{q0004349@sjtu.edu.cn}\\
  \And
  Duy-Tung Pham\\
  Khoury College of Computer Sciences\\
  Northeastern University, USA\\
  \texttt{pham.duy@northeastern.edu} \\
  \And
  Hoang-Son Do\\
  Institute of Mathematics\\
  Vietnam Academy of Science and Technology\\
  \texttt{dhson@math.ac.vn}\\
  University of Science and Technology of Hanoi\\
  Vietnam Academy of Science and Technology\\
  \texttt{do-hoang.son@usth.edu.vn}\\
  \And
  Tan Minh Nguyen$^\dagger$\\
  Department of Mathematics\\
  National University of Singapore\\
  \texttt{tanmn@nus.edu.sg}\\
  \And
  Thieu N. Vo$^\dagger$\\
  Department of Computer Science\\
  University of Bath, United Kingdom\\
  \texttt{ntv22@bath.ac.uk}
}

\begin{document}

\maketitle
{\def\thefootnote{$\dagger$}\footnotetext{Co-last authors. Correspondence to: \texttt{q0004349@sjtu.edu.cn} \& \texttt{ntv22@bath.ac.uk}.}}

\begin{abstract}
  We partially resolve Conjecture~1 of \textit{Karagodin~et~al., Clustering in Causal Attention Masking (NeurIPS~2024)} and extend it to time-dependent parameter regimes. The conjecture predicts that causal attention dynamics converge to two clusters aligned with the eigenvectors associated with the largest eigenvalue $\lambda_{\max}$ of the value transformation. We prove that this behavior holds when $\lambda_{\max}>0$ and is generally false when $\lambda_{\max}\le0$. The resulting two-cluster regime is common in practice and substantially more challenging than the classical single-cluster setting studied previously. A central contribution of our work is the identification of a rigorous dimension reduction phenomenon: despite evolving in a high-dimensional embedding space, token trajectories collapse onto a two-dimensional structure governed by a small number of dominant directions. Building on this insight, we introduce a new analytical framework--instability within the flow--inspired by instability phenomena in physical systems. This framework enables the analysis of steady-regime behavior in non-monotone, non-autonomous attention dynamics. Our theoretical findings are illustrated through simulations and experimental studies using  Group Query Attention (GQA) layers.
\end{abstract}

\section{Introduction}\label{sec:introduction}
    % {\color{red} ...to be edited later...}

    Transformer architectures \cite{vaswani2017attention} underpin modern artificial intelligence, achieving strong performance in natural language processing \cite{dehghani2018universal, dai2019transformer, al2019character, baevski2018adaptive,nielsen2024elliptical,raffel2020exploring}, computer vision \cite{dosovitskiy2020image, ramesh2021zero, radford2021learning, liu2021swin,abdullaev2025revisiting}, general machine learning \cite{tay2022efficient, khan2022transformers,nguyen2022fourierformer,lin2022survey}, and reinforcement learning \cite{janner2021offline, chen2021decision}. Their success is driven by large-scale pretraining, enabling effective transfer to downstream tasks \cite{radford2019language, devlin2018bert, radford2018improving,teo2025molex}.
    By replacing recurrence and convolution with self-attention, Transformers model global token interactions and capture long-range dependencies, yielding rich contextual representations \cite{lin2017structured, vig2019analyzing, cho2014learning, tenney2019bert,nguyen2022improvingkeys,nguyen2022head, parikh2016decomposable}. As a result, Transformers form the core architecture of large language models.
    Despite their empirical success, a rigorous theoretical understanding of their internal representations remains limited, an issue that becomes increasingly critical as these models continue to scale.

    Transformers can be interpreted through a particle-based perspective, where individual tokens are viewed as interacting entities. Adopting this viewpoint, the works of \cite{geshkovski2023mathematical,geshkovski2024emergence} investigate the continuous-time limit of Transformer models, reformulating their behavior as a dynamical system. Within this framework, the authors demonstrate that token representations evolve toward clustered configurations associated with specific attractors determined by their initial conditions, providing a theoretical explanation for the contextual nature of Transformer representations. 
    Further extending this line of analysis, \cite{geshkovski2024dynamic} shows that the self-attention dynamics can exhibit metastability, with tokens remaining confined near multi-cluster configurations over exponentially long time horizons. Complementary results are obtained in \cite{alcalde2024clustering}, where a hardmax, attention-only Transformer is examined in a discrete-time setting using similar analytical techniques.
    Furthermore, the authors in \cite{pham2025dynamical} analyze the flow of tokens across attention layers and investigate how different forms of positional encoding - specifically absolute and rotary - affect these dynamical regimes.

    Recent work on token dynamics has been extended to causal attention to better reflect its central role in sequential generation. In this autoregressive setting, each token is allowed to attend only to past tokens, enforcing temporal consistency by masking future information during attention computation. This restriction distinguishes causal attention from full attention, which operates on complete sequences and is therefore better suited to tasks such as machine translation.
    Because of its ability to support step-by-step generation, causal attention is widely used in large-scale generative models, including natural language and image generation systems such as DALL-E \cite{ramesh2021zero}, VQGAN \cite{esser2021taming}, and Parti \cite{yu2022scaling}, as well as multimodal foundation models like Chameleon \cite{team2024chameleon}. More generally, attention mechanisms that limit interactions to subsets of tokens have enabled recent scaling advances, leading to models such as MUSE \cite{chang2023muse} and AlphaFold~3 \cite{abramson2024accurate}.
    Unlike full attention, the dynamics induced by causal attention cannot be characterized as a mean-field gradient flow as in \cite{geshkovski2024dynamic, geshkovski2024emergence, geshkovski2023mathematical}, necessitating different analytical tools.

    % The conjecture by Karagodin et al., its importance, and link to papers cite to it....

    % \begin{table*}[ht]
    % \centering
    % \small
    % \begin{tabular}{lcccp{3.2cm}}
    % \toprule
    % \textbf{Work} & \textbf{Parameters} & \textbf{Initial Data} & \textbf{Key Techniques} & \textbf{Results} \\
    % \midrule
    % \cite{karagodin2024clustering}& $V$ constant & Arbitrary & Explicit $x_1(t)$ formula; & Convergence of $x_1(t)$ to $\pm \xi_{\max}$; \\
    %  & $Q,K$ general & on $\mathbb{S}^{d-1}$ & Schub genericity theorem & Full convergence  if $V=I_d$; \\
    %  \\
    % \midrule
    % \cite{abellaconsensus} & $V(t)$ symmetric& Single & Lyapunov function; & Global convergence to \\
    % & $Q(t),K(t)$ general & hemisphere & Input-to-state stability  & single cluster $\xi_{\max}$ \\
    % \midrule
    % Our work & $V(t)$ normal & Arbitrary & Instability analysis; & Global convergence to \\
    % & $Q(t),K(t)$ general & on $\mathbb{S}^{d-1}$ & Hierarchy dimension deduction & \textbf{two clusters $\pm\xi_{\max}$}; \\
    % & &  & Error-to-traverse ratio & Theoretical justification of\\
    % & & & & \textbf{dimensional deduction} \\
    % \bottomrule
    % \end{tabular}
    % \caption{\small Comparison of theoretical analyses for causal-attention dynamics. Our work resolves the two-cluster regime for diagonal $V(t)$ without hemisphere constraints and provides the first theoretical justification for the dimensional deduction phenomenon conjectured from experiments.}
    % \label{tab:prior_work_comparison}
    % \end{table*}

    \begin{table*}[ht]
    \centering
    \small
    \begin{tabular}{lllll}
    \toprule
    \textbf{Work} & \textbf{Parameters} & \textbf{Initial Data} & \textbf{Key Techniques} & \textbf{Theoretical Results} \\
    \midrule
    \cite{karagodin2024clustering}& $V$ constant & Arbitrary & Explicit $x_1(t)$ formula; & Convergence of $x_1(t)$ to $\pm \xi_{\max}$ \\
     & $Q,K$ general & on $\mathbb{S}^{d-1}$ & Schub genericity theorem & Full convergence  if $V=I_d$ \\
    \midrule
    \cite{abellaconsensus} & $V(t)$ symmetric& Single & Lyapunov function; & Global convergence to \\
    & $Q(t),K(t)$ general & hemisphere & Input-to-state stability  & single cluster $\xi_{\max}$ \\
    \midrule
    Ours & $V(t)$ symmetric & Arbitrary & Instability analysis; & Global convergence to \\
    & $Q(t),K(t)$ general & on $\mathbb{S}^{d-1}$ & Hierarchy dim. deduction & \textbf{two clusters $\pm\xi_{\max}$}; \\
    & &  & Error-to-traverse ratio & Theoretical justification of\\
    & & & & \textbf{dimensional deduction} \\
    \bottomrule
    \end{tabular}
    \caption{\small Comparison of theoretical analyses for causal-attention dynamics. Our work resolves the two-cluster regime for diagonal $V(t)$ without hemisphere constraints and provides the first theoretical justification for the dimensional deduction phenomenon conjectured from experiments.}
    %\vspace{-0.8cm}
    \label{tab:prior_work_comparison}
    \end{table*}

   \subsection{Background -- Causal Attention Dynamics and a Two-Clustering Conjecture}
    
    A full Transformer layer typically consists of multi-head self-attention, a position-wise feed-forward network (FFN), normalization layers, and residual connections~\citep{vaswani2017attention}. Our goal is not to reproduce every component of the Transformer architecture, but rather to clarify and formalize the key mathematical structure that governs its behavior. To this end, we simplify the architecture to only the components relevant for our analysis: a single-head self-attention layer with normalization and a residual connection, omitting the FFN block.
    
    Let the input sequence be represented by token embeddings
    \(X = [x_1,\ldots,x_L] \in \mathbb{R}^{L \times d}\), where \(L\) is the number of tokens and \(d\) is
    the embedding dimension.  
    With parameter matrices
    \(Q,K,V \in \mathbb{R}^{D \times D}\) used to form queries, keys and
    values, the attention operator is given by
    % \[
    % \operatorname{Attn}(X)
    %   =\operatorname{softmax} \left( \frac{(XQ) (XK)^{\top}}{\sqrt{D}} \right) \cdot XV,
    % \]
    % or equivalently
    \begin{align*}
        \operatorname{Attn}(X)_i = \sum_{j=1}^L \left( \frac{e^{\left< Qx_i,Kx_j \right>}}{\sum_{k=1}^Le^{\left< Qx_i,Kx_j \right>}}  \right) Vx_j,
    \end{align*}
    for $i=1,\ldots,L$.
    
    Following \cite{zhang2019root}, we add an RMS normalization layer and a residual connection.  
    Denoting by \(X^t\) the activations at layer \(t\), the update to the next layer is
    \begin{align}\label{eq:single_head_update}
        X^{t+1} = X^{t} + \operatorname{RMSNorm}(\operatorname{Attn}(X^{t})).
    \end{align}
    The update rule~\eqref{eq:single_head_update} can be interpreted as a finite-difference approximation of a time derivative, with \(X^{t+1}-X^{t}\) playing the role of the increment between consecutive layers, and the RMS normalization playing the role of keeping the token representations lie on the standard unit sphere \(\mathbb{S}^{D-1}\).
    Hence, the sequence \(\{X^{t}\}_{t}\) can be regarded as a time-discretization of an underlying continuous-time flow which is given by the dynamical system
    \begin{align*}%\hspace{-0.3cm}
        \pt_t x_i(t)  = \mathbf{P}_{x_i(t)} \Big( \sum_{j=1}^L a_{ij}(t)  V(t)x_j(t) \Big),
    \end{align*}
    with
    $$a_{ij}(t) = \frac{e^{\left< Q(t)x_i(t),K(t)x_j(t) \right>}}{\sum_{k=1}^Le^{\left< Q(t)x_i(t),K(t)x_j(t) \right>}},$$
    and the operator
    \(\mathbf{P}_{x}(y) := y - \langle x,y\rangle x / \lVert x\rVert^{2}\) is the projector
    that ensures \(x_{k}(t)\) remains on the sphere. 
    Finally, by imposing a causal masking constraint, so that each token
    attends only to its predecessors, we obtain the following causal self-attention equation that we analyze in this work:
    \begin{align}\label{eq:causal_dynamic_1}
        \pt_t x_i(t) = \mathbf{P}_{x_i(t)} \Big( \sum_{j=1}^i a_{ij}(t) V(t)x_j(t) \Big),
    \end{align}
    with
    $$a_{ij}(t) = \frac{e^{\left< Q(t)x_i(t),K(t)x_j(t) \right>}}{\sum_{k=1}^ie^{\left< Q(t)x_i(t),K(t)x_j(t) \right>}},$$
    for $i,j=1,\ldots,L$.

    The authors in \cite{karagodin2024clustering} predicted that the eigenvectors of the value matrix plays a central role in determining the qualitative clustering behavior of the causal attention dynamics. In the time-independent setting where $Q$, $K$, and $V$ are fixed, the authors in \cite{karagodin2024clustering} further conjectured the following.

    \begin{conjecture}[Two-clustering configuration of the causal attention dynamics]\label{conj:two-clustering}
    Assume that $Q,K,V$ are arbitrary time-independent matrices and that $V$ has $d$ distinct positive eigenvalues. Let $\lambda_{\max}$ denote the largest eigenvalue, and let $\xi_{\max}$ be a corresponding unit eigenvector, i.e., $|\xi_{\max}|=1$. Then, for almost every initial condition
    \[
    X(0)=(x_1(0),\ldots,x_L(0)) \in (\mathbb{S}^{d-1})^L,
    \]
    the causal attention dynamics~\eqref{eq:causal_dynamic_1} converges to a two-cluster configuration:
    \[
    \lim_{t\to+\infty} x_i(t) \in \{-\xi_{\max},\xi_{\max}\}, \qquad \text{for all } i=1,\ldots,L.
    \]
    \end{conjecture}
    
    This two-cluster behavior has been observed empirically for causal attention in \cite{karagodin2024clustering} and for full attention in \cite{pham2025dynamical}. To the best of our knowledge, however, a rigorous proof of this conjecture is not currently available in the literature.

    \subsection{Our main contributions}

    % {\color{red} ...need to shorten this subsection...}
    
    In this paper, we prove Conjecture~\ref{conj:two-clustering} under a mild condition and further generalize it to settings with time-dependent parameters $Q(t), K(t)$ and $V(t)$. 
    % A comparison of our results with related ones in the existing literature is provided in Table~\ref{tab:prior_work_comparison}. 
    Our rigorous theoretical analysis reveals several dynamical mechanisms that govern when and how two-cluster configurations emerge, including hierarchical suppression, intrinsic dimensional reduction, and recurrence to the dominant directions~$\pm\xi_{\max}$.

    \begin{itemize}
    
    \item \textbf{Recurrence to the dominant direction $\xi_{\max}$.}
    % Even with depth--dependent attention and value matrices, dominance is not lost.
    Once the first $L-1$ tokens stabilize, the last token is forced to repeatedly
    re-align with  $\xi_{\max}$ along arbitrarily large depths (Theorem~\ref{thm:two-dim-effective}). 
    % Apart from a single exceptional trajectory in the reduced two--dimensional dynamics, all token paths revisit the dominant state. 
    This recurrent re-alignment creates a ``déjà vu'' effect: past dominant information is not permanently forgotten, but reappears again and again in the infinite--depth limit.

    \item \textbf{Hierarchical suppression mechanism.} %and error-to-transverse ratio.} 
    % We introduce the \emph{error-to-transverse ratio} (see~\eqref{ratio}), a simple but crucial quantity that captures the competition between two forces in causal attention dynamics: the residual signal carried forward by earlier tokens and the geometric scale of the transverse component of the current token. 
    % This ratio turns out to govern a striking \emph{hierarchical suppression}: depending on its size, the system either collapses entire blocks of coordinates or locks onto exactly one surviving direction (see Theorem~\ref{sup-main}). 
    We introduce the \emph{error-to-transverse ratio} $\kappa_\ell(t)$ (see definition in~Eq.~\eqref{eq:kappa}), a simple but crucial quantity that governs a striking \emph{hierarchical suppression}: depending on its size, the system either collapses entire blocks of coordinates or locks onto exactly one surviving direction (see Theorem~\ref{thm:formal-A6}).
    % To the best of our knowledge, no prior work has identified, quantified, or analyzed this competition between upstream error and transverse geometry. 
    The error-to-transverse ratio  exactly predicts when a token will suppress all but its most  favorable direction.
    
    \item \textbf{Dimensional reduction.} 
    The token dynamics concentrates on a low-dimensional manifold, revealing \emph{dimensionality reduction} within deep sequence models. 
    % In the time-independent setting, we show that whenever the Dominant-Direction Selection Condition (DDSC) holds—meaning that the residual error from earlier tokens becomes negligible compared to the transverse magnitude of the current token, while at least one transverse coordinate remains nonvanishing—the nonlinear $d$-dimensional dynamics collapse onto a \emph{finite} set of admissible limit states (Theorem~\ref{generalQK}), each of which can then be analyzed explicitly through its linearized eigenvalues (see Lemma~\ref{upper}). Under DDSC, 
    Under mild conditions, every nontrivial limiting configuration lies in a two-dimensional coordinate plane and satisfies a single scalar algebraic constraint (Theorem~\ref{thm:generalQK_main}).
    This rigidity phenomenon appears to be new and, to our knowledge, has not been rigorously analyzed for attention dynamics in the literature.
    \end{itemize}

Existing analyses of the causal-attention ODE can be categorized by their assumptions on the parameter matrices $(V, Q, K)$ and initial data. 
A summary of the theoretical landscape and the key analytical ideas is provided in Table~\ref{tab:prior_work_comparison}.
Our theoretical findings are illustrated through simulations and experimental studies using  Group Query Attention (GQA) layers.
% Our findings are carefully illustrated through simulations and experimental studies using GPT-3 models.

\paragraph{Difficulties from previous  approaches.}
% The two-clustering case is far more complicated than the single clustering one which has been analyzed in \cite{karagodin2024clustering} and \cite{abellaconsensus}. 
The two-cluster regime with time-varying depth is far more challenging than the classical single-cluster setting, which was the primary focus of the analyses in \cite{karagodin2024clustering} and \cite{abellaconsensus}.
The dynamics no longer admit a monotone Lyapunov functional, and thus every existing technique in the literature -- 
including hemisphere-based Lyapunov methods~\cite{abellaconsensus} and Schub-type genericity arguments~\cite{karagodin2024clustering} --
completely fails in this regime.  

% {\color{red} to be moved to other place:} In the time-dependent setting, the dynamics become non-autonomous, a regime that is notoriously difficult to analyze in the absence of any monotone structure or Lyapunov functional. In fluid dynamics, for example, the standard linear-to-nonlinear frameworks -- such as Grenier’s nonlinear iteration scheme {\color{red}\cite{????}} or Strauss’s classical instability theory for ideal plane flows {\color{red}\cite{????}} -- apply only to time-independent linearized operators. Once the linearization depends explicitly on time, the available theory becomes extremely limited: there is, in general, no mechanism guaranteeing stability, persistence of structure, or convergence for non-autonomous flows. \red{to do: please comment on why theory/theorem is limited for time-dependent}) From a dynamical-systems viewpoint, tools such as Shub-type structural stability theorems rely crucially on an autonomous vector field and have no direct analogue in the time-dependent case, where recurrent or even chaotic behavior is often the most one can expect.  In this work, we take on this problem using the full range of new analytical tools and structural insights developed throughout the paper, allowing us to penetrate a regime that has long remained beyond reach.

\paragraph{Our new approach.}
Our key contribution is a new mathematical tool that we call the \emph{instability-within-the-flow}, inspired by stability theory in physics. The evolution of a token under a time-dependent value matrix $V(t)$ resembles a small ball rolling on a hill whose summit $r(t)$ itself moves over time. As the ``hilltop'' $r(t)$ drifts, even slowly, the stable and unstable regions experienced by the token shift accordingly, producing a much richer and more delicate dynamical landscape. A rapidly moving hilltop can trap the flow in an interior region (leading to interior clusters), while a stationary hill forces the trajectory toward one of two extremal states $\{\xi_{\max},-\xi_{\max}\}$. To our knowledge, this structure has not appeared in previous work, and we believe it offers a promising mathematical tool for understanding directional biases and cluster formation in deep sequence models. The full framework is developed in Sections~\ref{sec:moving_ball} in the main text and in Section~\ref{instability} in appendix.

\textbf{Organization.} Section~\ref{sec:dim_reduction} analyzes hierarchical suppression and high-dimensional dimension-reduction mechanisms. Section~\ref{sec:moving_ball} visualizes causal attention dynamics as the motion of a small ball on a hill. Section~\ref{sec:dominant_direction} proves the two-cluster behavior of tokens. Section~\ref{sec:experiment} validates the theoretical findings in practical Transformers with GQA layers.

% \section{Preliminaries}

% In this section, we recall the original Transformer and present a simplified model for the attention dynamic that we will consider in throughout this paper.

\section{Hierarchical Suppression Mechanism and Dimension Reduction}\label{sec:dim_reduction}

Our first main result is to uncover a high-dimensional reduction mechanism in which the dynamics collapse to exactly two coordinates for arbitrary matrices \(Q(t)\) and \(K(t)\).
Although the dynamics formally reside in a high-dimensional space, the long-time behavior is effectively governed by only a handful of intrinsic degrees of freedom.

The study of the high-dimensional reduction mechanism based on the dynamical system~\eqref{eq:causal_dynamic_1} without any additional assumption on the model parameters $Q(t),K(t)$, and $V(t)$ is challenging. 
To make the proofs feasible while still maintaining the generality practicality of our theoretical findings, we will use the following assumption during this section (see also in \cite{abellaconsensus}).

\begin{assumption}[Symmetric and time-invariant eigenspace value matrix]\label{assum:V-symmetric}
    The value matrix $V(t)$ is a symmetric matrix having $d$ different eigenvalues and time-invariant eigenspaces.
    Furthemore, by a suitable linear change of variables, we can write without loss of generality that:
    \[
    V(t)=\diag\{\lambda_1(t),\lambda_2(t),\dots,\lambda_d(t)\}, 
    \]
    so that the eigen-basis coincides with the canonical orthogonal basis $\{\vec e_1,\dots,\vec e_d\}$ 
    and 
    \begin{align*}
        \lambda_1(t)>\lambda_2(t)>\cdots>\lambda_d(t)\quad\text{for all}\quad t\ge 0.
    \end{align*}
\end{assumption}
% assume throughout this paper that the value matrix $V(t)$ is a symmetric matrix having $d$ different eigenvalues and time-invariant eigenspaces.
% Under Assumption~\ref{assum:V-symmetric}, by a suitable linear change of variables, we can write without loss of generality that:
% \[
% V(t)=\diag\{\lambda_1(t),\lambda_2(t),\dots,\lambda_d(t)\}, 
% \]
% so that the eigen-basis coincides with the canonical orthogonal basis $\{\vec e_1,\dots,\vec e_d\}$ 
% and 
% \begin{align*}
%     \lambda_1(t)>\lambda_2(t)>\cdots>\lambda_d(t)\quad\text{for all}\quad t\ge 0.
% \end{align*}
Under Assumption~\ref{assum:V-symmetric}, the largest eigenvalue $\vec e_1=\xi_{\max}$ will be the leading direction selected by the dynamics.
% \beq\label{order-V}
% \lambda_1(t)>\lambda_2(t)>\cdots>\lambda_d(t)\quad\text{for all}\quad t\ge 0.
% \eeq
Let us fix a solution $\mathbf{x}(t)=(x_1(t),\ldots,x_L(t)) \in C^{\infty}([0,+\infty),\mathbb{S}^{d-1})^L$ of the dynamical system~\eqref{eq:causal_dynamic_1}, subject to the initial condition $\mathbf{x}(0) \in (\mathbb{S}^{d-1})^L$.
We will derive our theorems by induction on the number $L$ of tokens.
In particular, we define the error function
\[
\eps_{L-1}(t)=\sum_{j=1}^{L-1} \|x_j(t) - s_j \vec e_1 \|, 
\]
where $\|\cdot\|$ denotes the Euclidean norm.
 Our inductive hypothesis for the first $L-1$ tokens is
\begin{equation}\label{induction}
\eps_{L-1}(t)\;\longrightarrow\; 0
\qquad \text{as } t \to \infty,
\end{equation}
Here $s_j$ simply records which cluster the token $x_j$ converges to: $+1$ 
corresponds to the ``top'' copy of $\vec e_1$ and $-1$ to the ``bottom'' copy 
$-\vec e_1$. 
Since we are now focusing on the dynamic of the $L$-th token, to simplify the notation, we will denote 
\begin{align}\label{eq:y}
    y(t)=(y_1(t),\ldots,y_d(t)):=x_L(t).
\end{align}
Here, $y_1$ measures alignment with the dominant direction, and the remaining
coordinates $(y_2,\dots,y_d)$ represent transverse components.

Our goal in this section is to identify the leading dynamics of $y$. 
We first prove that the dynamic of $y(t)$ induce a hierarchical suppression mechanism as follows:
% See Theorem~\ref{sup-main} for a precise and fully rigorous formulation.

% We work in a coordinate system adapted to the dominant direction of the dynamics.
% Specifically, we fix an orthonormal basis $\{\vec e_1,\dots,\vec e_d\}$ such that
% $\vec e_1=\xi_{\max}$, the leading direction selected by the dynamics.
% The token state $y(t)\in\mathbb S^{d-1}$ is then expressed in coordinates
% \[
% y(t) = (y_1(t),\dots,y_d(t)),
% \]
% where $y_1$ measures alignment with the dominant direction, and the remaining
% coordinates $(y_2,\dots,y_d)$ represent transverse components.

\begin{theorem}[hierarchical suppression mechanism -- See Theorem~\ref{sup-main}]\label{thm:formal-A6}
% Let $y(t)\in\mathbb S^{d-1}$ denote the $L$-th token, expressed in an orthonormal
% coordinate system $\{e_1,\dots,e_d\}$ with $e_1=\xi_{\max}$.
% Let $\varepsilon_{L-1}(t)\ge0$ be the residual error contributed by the first $L-1$ tokens.
For each level $\ell\in\{2,\dots,d\}$, define
\begin{align}\label{eq:kappa}
    R_\ell(t)=\Bigl(\sum_{j=\ell}^d y_j(t)^2\Bigr)^{1/2},
    \quad \text{and} \quad
    \kappa_\ell(t)=\frac{\varepsilon_{L-1}(t)}{R_\ell(t)}.
\end{align}

Assume that for some $\ell\ge2$,
\beq\label{DDSC-intro}
\kappa_\ell(t)\longrightarrow 0
\qquad\text{as } t\to\infty.
\eeq
Then the dynamics enforce a hierarchical suppression among the transverse coordinates
$\{y_\ell,\dots,y_d\}$ in the following sense: exactly one of the two alternatives holds:

\begin{enumerate}
\item (\emph{Dominant transverse direction})  
There exists an index $k\in\{\ell,\dots,d\}$ such that
\[
y_j(t)\longrightarrow 0
\qquad\text{for all } j\notin\{1,k\},
\]
and the trajectory converges to the invariant two-dimensional subspace spanned by
$\{e_1,e_k\}$.

\item (\emph{Recursive suppression})  
The coordinate $y_\ell(t)\longrightarrow 0$, while the remaining transverse vector
$(y_{\ell+1}(t),\dots,y_d(t))$ remains nontrivial. In this case the same mechanism
applies at the next level $\ell+1$, with $\kappa_{\ell+1}(t)$ governing the subsequent
suppression.
\end{enumerate}

Iterating this reduction yields a unique surviving transverse coordinate $y_k$ for some
$k\ge2$, and the trajectory collapses onto a low-dimensional invariant subsystem spanned by
$(y_1,y_2,\dots,y_{\ell-1},y_k)$.
\end{theorem}

Roughly speaking, by the above theorem, we uncover a hierarchical suppression mechanism acting on the transverse coordinates:
As the residual error contributed by earlier tokens becomes negligible relative to the remaining transverse energy, the system progressively eliminates all but one transverse direction, yielding an effective reduction in dimensionality.

Our key idea for the proof of this theorem is to separate the direction dynamics from the magnitude dynamics: once we factor out the overall magnitude of a block of coordinates, the remaining dynamics describe a competition over which direction the token points to.
Because different coordinates have systematically different growth advantages (encoded by the ordered of the eigenvalues of the value matrix $V$), this directional dynamics has a built-in drift that continuously sharpens the representation, pushing it toward the boundary where some coordinates vanish.
As a result, the system cannot remain in a persistent mixed state: either a given coordinate loses mass and is suppressed, or it survives and eventually dominates, eliminating all weaker competitors.
The only subtlety is noise inherited from earlier tokens, but as long as this residual signal decays faster than the scale of the current competition, it cannot prevent concentration.
In effect, the dynamics behave like an expert-selection process: coordinates are tested one by one, and any direction that remains consistently active must ultimately win decisively.
This explains why, despite the ambient dimension, the token collapses onto a low-dimensional structure driven by a small number of dominant features. 
The full detail of the proof of Theorem~\ref{thm:formal-A6} is presented in Section~\ref{reduce-2d}.

Regarding the dimension reduction mechanism, we further prove another high-dimensional reduction mechanism in which the dynamics collapse to exactly two coordinates, as we see in the following theorem.

\begin{theorem}[Dimension reduction mechanism -- See Theorem \ref{generalQK}]\label{thm:generalQK_main}
Let $Q$ and $K$ be arbitrary matrices.  
Assume that $x_i(t)\to \pm \vec e_1$ for all $i\le L-1$, and that the  
condition~\eqref{DDSC-intro} holds for the final $L^{th}$ token.  
Then there exists a finite subset $\mathcal{S}\subset\mathbb{S}^{d-1}$ such that,
for almost every initial condition, the trajectory $x_L(t)$ converges to some 
$y_\star\in\mathcal{S}$ as $t\to\infty$.

Moreover, if the limit $y_\star$ is not equal to $\pm\vec e_1$, then $y_\star$ 
must lie in a two–dimensional coordinate plane and take the form
\[
(\alpha_1,0,\dots,0,\al_\ell,0,\dots,0),
\qquad \alpha_1^2+\al_\ell^2=1,
\]
that is, only the first and the $\ell$-th coordinates are nonzero. 
\end{theorem}

By Theorem~\ref{thm:generalQK_main}, we uncover a dimensional reduction mechanism in which the dynamics collapse to exactly two coordinates for arbitrary matrices \(Q\) and \(K\).
Intuitively speaking, once the parameter matrices $Q$ and $K$ have stabilized, we shows that, asymptotically, the token dynamics are confined to the one-dimensional manifold
$\mathbb{S}^1 \times \{0\}^{d-2}$.
This residual degree of freedom corresponds to a phase (or orientation) ambiguity in the embedding space. A natural question is whether the token may continue to rotate along this manifold indefinitely, even after the training dynamics appear to have converged. Using tools from complex analysis, we prove that the set of admissible long-time limit points on $\mathbb{S}^1$ is necessarily discrete. Consequently, the token representation must converge to a single fixed configuration (see Lemma~\ref{son-lem-3}). In particular, apparent stabilization of the parameter matrices $Q$ and $K$ implies convergence of the token dynamics. We refer to Theorem~\ref{generalQK} and Section~\ref{proofQK} for the complete argument.

\section{Visualization of Tokens' Evolution as Motion of a Small Ball on a Hill}\label{sec:moving_ball}

The self-attention mechanism--which updates token representations based on a weighted combination of other tokens--lies at the heart of their expressive power. While recent theoretical work has analyzed the equilibrium states of such systems under static attention patterns, real training dynamics involve rapidly shifting attention, raising a fundamental question: \emph{How do token embeddings evolve when the attention landscape itself changes over time?}
In this section, we propose another dynamical perspective that captures this phenomenon intuitively from the motion of a small ball on a hill. 

\paragraph{Moving hills and metastable clustering.}
% We propose a dynamical perspective that captures this phenomenon intuitively. 
Consider a token's trajectory under a time-dependent attention matrix $V(t)$ as the motion of a ball on a hill whose summit itself moves. 
In the static case, the ball inevitably rolls to one of two stable fixed points (analogous to token collapse to $\pm \xi_{\max}$). 
But if the hilltop moves sufficiently fast, the ball may become trapped near an intermediate position, giving rise to interior clusters--token representations that do not converge to extremal points but instead settle into a metastable or oscillatory state.

To make this intuition concrete, we consider a minimal scalar model:
\[
x'(t) = (1-x^2)(x - r(t)), \qquad x(t) \in (-1,1).
\]
If $r(t) \equiv r_0$ is constant, almost every initial condition converges to $\pm 1$, unless initialized exactly at the unstable equilibrium $x = r_0$.  
But if $r(t)$ moves rapidly between $-1$ and $1$, the ball cannot roll off either side; it remains trapped near an intermediate value, forming an interior cluster.

\paragraph{A tractable attention model.}
We now connect this toy analogy to a simplified Transformer setting.  
Assume $Q^\top K = I_d$ and fix the first token as $x_1(t) \equiv \vec e_1$.  
Let the second token be $x_2(t) = x(t)\vec e_1 + y(t)\vec e_2$.  
Its dynamics reduce (see Theorem~\ref{toy-case}) to a scalar equation of the form
\[
x'(t) = (1-x^2) \, h(x) \, g(t,x),
\]
where $g(t,x)$ encodes the time-varying attention from the first token, and $h(x) > 0$ is a positive weighting.  
Crucially, under mild conditions on the attention coefficients $\lambda_1(t)$ and $\lambda_2(t)$, there exists a moving root $r(t) \in (-1,1)$ satisfying $g(t,r(t)) = 0$, which plays the role of the moving hilltop.  
If $\lambda_2(t) \to \lambda_{2,\infty} \in (0,1)$, then $r(t) \to r_\infty \in (-1,0)$, and the interior cluster stabilizes.

\paragraph{Deneric convergence to extremal or metastable states.}
Our main technical contribution in this section is a rigorous treatment of this moving-root mechanism in the presence of small perturbations.  
We prove that, under natural regularity conditions, the dynamics admit at most one exceptional trajectory $x_\star(t)$ that does not converge to $\pm 1$.  
All other initial conditions are driven to the boundary, except possibly those starting exactly on this special interior path.
In particular, we prove that:

\begin{theorem}[See Theorem~\ref{almost}]\label{thm:almost_main}
Consider the perturbed dynamics
\[
x'(t) = (1-x^2) \, h(x) \, g(t,x) + w(t,x),
\]
where $w(t,x)$ is a small, asymptotically vanishing perturbation.  
Assume that $h(x) > 0$, and $\partial_x g(t,x) \ge c_0 > 0$, as well as there exists a moving root $r(t) \in (-1,1)$ with $g(t,r(t)) = 0$ and $r'(t) \to 0$.  
Then:
\begin{enumerate}
    \item There exists at most one interior trajectory $x_\star(t)$ that avoids convergence to $\pm 1$.
    \item If $\sup_x |w(t,x)| \to 0$, then for \emph{almost every} initial condition, we have $x(t) \to \pm 1$.
\end{enumerate}
\end{theorem}

Intuitively speaking, this result formalizes the hilltop analogy: a rapidly moving but asymptotically slowing root $r(t)$ can sustain at most one persistent interior path; all other trajectories collapse to the poles.  
The analysis provides a mathematical foundation for understanding how time-varying attention can produce metastable token structures, a phenomenon observed in training dynamics but previously lacking a rigorous dynamical explanation.

Theorem~\ref{thm:almost_main} is then used to analyze the dominant direction of the dynamic of the leading token $y(t)$ (see equation~\eqref{eq:y}).
Indeed, under suitable block-structure assumptions on $Q(t)$ and $K(t)$, we derive an effective differential equation governing the evolution of the projection $x_L \cdot \xi_{\max} = y_1$, treating all remaining coordinates as perturbations. This reduction is carried out in Theorem~\ref{all-zero}. Formally, the dynamics of $y_1$ take the form
\begin{equation}\label{y1-dyn_main}
\partial_t y_1 = (1 - y_1^2)\, M(t,y_1) + o(1),
\end{equation}
where $o(1)$ denotes a quantity that vanishes as $t \to \infty$. Since the equation is non-autonomous and the right-hand side is not monotone for the  two-cluster regime $\{\xi_{\max}, -\xi_{\max}\}$, we cannot linearize around a stationary solution and apply the stable manifold theorem as in \cite{karagodin2024clustering,geshkovski2023mathematical}, nor use the Lyapunov method as in \cite{alcalde2024clustering}. Instead, the right-hand side of \eqref{y1-dyn_main} admits a  ``moving equilibrium'' in the interior of $(-1,1)$--a phenomenon already visible in the simpler setting \eqref{toy-1}. This behavior corresponds precisely to the moving-hilltop analogy in the stability theory of physical systems, which we discuss in Section~\ref{instability}. Our key insight is that, under suitable time-varying assumptions on the parameter matrices, this unstable equilibrium cannot trap the solution $y_1$ in the interior of $(-1,1)$. Consequently, $y(t)$ must converge to one of the two clusters: $\xi_{\max}$ (corresponding to $y_1 = 1$) or $-\xi_{\max}$ (corresponding to $y_1 = -1$). The full analytical framework is developed in Theorem~\ref{almost} and Section~\ref{full-proof}. Importantly, our analysis is not limited to establishing convergence to the two-cluster configuration. Rather, it also provides insight into alternative clustering behaviors that may arise in practice and elucidates the instability mechanisms that can unlock such dynamics. In particular, we present an example in which oscillatory drifting of the unstable equilibrium within the interior can trap the $L$-th token and lead to dynamics distinct from convergence to $\pm \xi_{\max}$. This demonstrates that our assumptions are essentially sharp; see Proposition~\ref{trap-case}.

Our work in this section bridges intuitive dynamical analogies with precise asymptotics, offering a new lens through which to study token evolution in Transformers.  
We believe this perspective will inform both theoretical analysis and the design of stable training procedures.

\section{Recurrent Collapse to the Dominant Direction}\label{sec:dominant_direction}

% In this section, we combine the results of the preceding sections to prove that, during training, the token representations cluster into two antipodal directions under certain scenario. 
% In particular, the dimensional-collapse results explain how high-dimensional trajectories simplify over time, while the recurrent-return property accounts for why most tokens ultimately align with the dominant eigen-direction.

Transformer training dynamics often exhibit a striking phenomenon: despite the high-dimensional embedding space, token representations tend to converge toward a small number of dominant directions. In this section, we combine the results of the preceding sections to formalize this behavior by showing that under slowly varying attention patterns, token trajectories ultimately reduce to two-dimensional dynamics, with nearly all solutions recurrently approaching the dominant direction determined by the value matrix.

\begin{assumption}[Block structure of the query-key matrix]\label{assum:dominant-direction}
In addition to Assumption~\ref{assum:V-symmetric}, we will also use the following assumptions in this section:
\begin{itemize}
    \item[($i$)] The query-key matrix $\mathcal T(t)=Q^\top K(t)$ has the block structure
    \begin{equation}\label{T-condition}
    \mathcal T(t)=
    \begin{pmatrix}
    m_0(t) & 0 \\
    0 & m_1(t) I_{d-1} + \mathcal B(t)
    \end{pmatrix},
    \qquad 
    \mathcal B(t)^\top=-\mathcal B(t),
    \end{equation}
    for some functions $m_0(t)>0$ such that
    $m_1(t)<m_0(t)+\frac{1}{2\beta}$ for all $t\ge0$.

    \item[($ii$)] Furthermore,
    $\lim_{t\to\infty}(V(t),\mathcal T(t))=(V_\infty,\mathcal T_\infty)$, and
    $\lim_{t\to\infty}V'(t)=0$,
    where
    \[
    V_\infty=\diag\{\lambda_{1,\infty},\ldots,\lambda_{d,\infty}\},
    \qquad
    \lambda_{1,\infty}>\lambda_{2,\infty}>\cdots>\lambda_{d,\infty},
    \qquad
    \lambda_{1,\infty}>0.
    \]
\end{itemize}
\end{assumption}
We note that the setting considered under Assumption~\ref{assum:dominant-direction} is substantially more general than the weight-tying framework studied in~\cite{karagodin2024clustering}. In particular, our formulation goes beyond weight tying by allowing the model parameters to depend on time. Under this assumption, we prove the following result.

\begin{theorem}[{See  Theorem~\ref{time-dependent-main} for a formal statement}] \label{thm:two-dim-effective}
    Assume that $x_1(t), \dots, x_{L-1}(t) \to \pm \xi_{\max}$, where $\xi_{\max}$ is the dominant eigen-direction of the limiting attention structure.
    Let us consider the evolution of a token $x_L(t) \in \mathbb{S}^{d-1}$ under self-attention with time-dependent weight matrices $V(t)$ and query-key product $\mathcal T(t) = Q(t)^\top K(t)$.  
    % Assume that:
    % \begin{enumerate}
    %     \item  $x_1(t), \dots, x_{L-1}(t) \to \pm \xi_{\max}$, where $\xi_{\max}$ is the dominant eigen-direction of the limiting attention structure.
    %     \item $\mathcal T(t)$ respects a block structure that decouples the dominant direction from the transverse subspace, with the transverse block strictly weaker than the dominant block.
    %    \item $V(t)$ and $\mathcal T(t)$ converge to fixed limits $V_\infty$ and $\mathcal T_\infty$, and the rate of change of attention decays ($V'(t) \to 0$). Here $V_\infty$ has maximum eigenvalue $\lambda_{\max}$ with corresponding eigenvector $\xi_{\max}$.
    % \end{enumerate}
    Then aside from a single exceptional trajectory, every solution \emph{recurrently approaches} the dominant state $\pm\xi_{\max}$.
    \[
    x_L(t_n)\to \pm \xi_{\max}\quad\text{along some subsequence}\quad t_n\to\infty.
    \]
\end{theorem}
% (Full technical statement and proof appear Theorem~\ref{time-dependent-main}.)

% Theorem~\ref{thm:two-dim-effective} provides a dynamical explanation for the empirically observed tendency of tokens to cluster into a small number of directions during training.  
% The dimensional collapse shows how high-dimensional trajectories simplify; the recurrent return property explains why most tokens end up aligned with the dominant eigen-direction.

% The coefficients $m_0(t)$ and $m_1(t)$ control the relative strength of the interaction along and perpendicular to $\vec e_1$, while $\mathcal B(t)$ contributes only rotational effects within the transverse space.

% \begin{remark}[{Relating to Conjecture~\ref{conj:two-clustering}}]
    
%     {\color{red} ...explain why Theorem~\ref{thm:almost_main} resolve Conjecture~\ref{conj:two-clustering}.}
% \end{remark}

We further prove that the condition is unavoidable $\lambda_{\text{max}}>0$ in the following theorem.

\begin{theorem}[Sharpness of the condition $\lambda_1> 0$] \label{thm:toy-case_main}
Suppose that $\lambda_1<0$. Then there exist $\beta>0$ and $\lambda_2<\lambda_1$ such that, for $L=d=2$, $Q=K=I_2$, $V=\operatorname{diag(\lambda_1,\lambda_2)}$, and $x_1(t)\equiv \vec e_1$, one can find a nonempty open set $U\subset \S^{1}$ with the property that for all $x_2(0)\in U$, the limit 
$\lim_{t\to\infty} x_2(t)$
exists and satisfies 
\[
\lim_{t\to\infty} x_2(t)\notin \{\pm \vec e_1\}.
\]
\end{theorem}

Intuitively speaking, Theorem~\ref{thm:two-dim-effective} shows that tokens tends to the unit length eigenvectors $\pm \xi_{\operatorname{max}}$ of the value matrix $V(t)$ whenever Assumption~\ref{assum:dominant-direction} holds.

\begin{remark}[Limitations of Assumption~\ref{assum:dominant-direction} in practical Transformers]\label{rem:limitation}
Assumption~\ref{assum:dominant-direction} is unlikely to hold exactly in practical pre-trained Transformers, since the value matrix $V(t)$ need not be symmetric or have time-invariant eigenvectors, and the key-query matrix $\mathcal T(t)$ need not possess the prescribed block structure.
Nevertheless, this assumption covers the weight-tying setting considered in \cite{karagodin2024clustering}, and therefore yields a proof of Conjecture~\ref{conj:two-clustering} under that setting.
Moreover, our formulation goes beyond the weight-tying case by allowing the model parameters to be time-dependent.
\end{remark}

% \vspace{-0.4cm}
\section{Empirical Validation}\label{sec:experiment}
% \vspace{-0.2cm}
In this section, we validate the emergence of the two-cluster regime through simulations with both single-head attention and in Grouped-Query Attention (GQA)~\citep{ainslie2023gqa}, a more general mechanism. We then investigate the role of the value projection matrix, highlighting its importance in shaping token dynamics and model adaptation.

% \vspace{-0.2cm}
\subsection{Validation on Grouped-Query Attention}

\begin{figure}[t]
    \centering
    \begin{subfigure}[t]{0.3\linewidth}
        \centering
        \includegraphics[width=\linewidth]{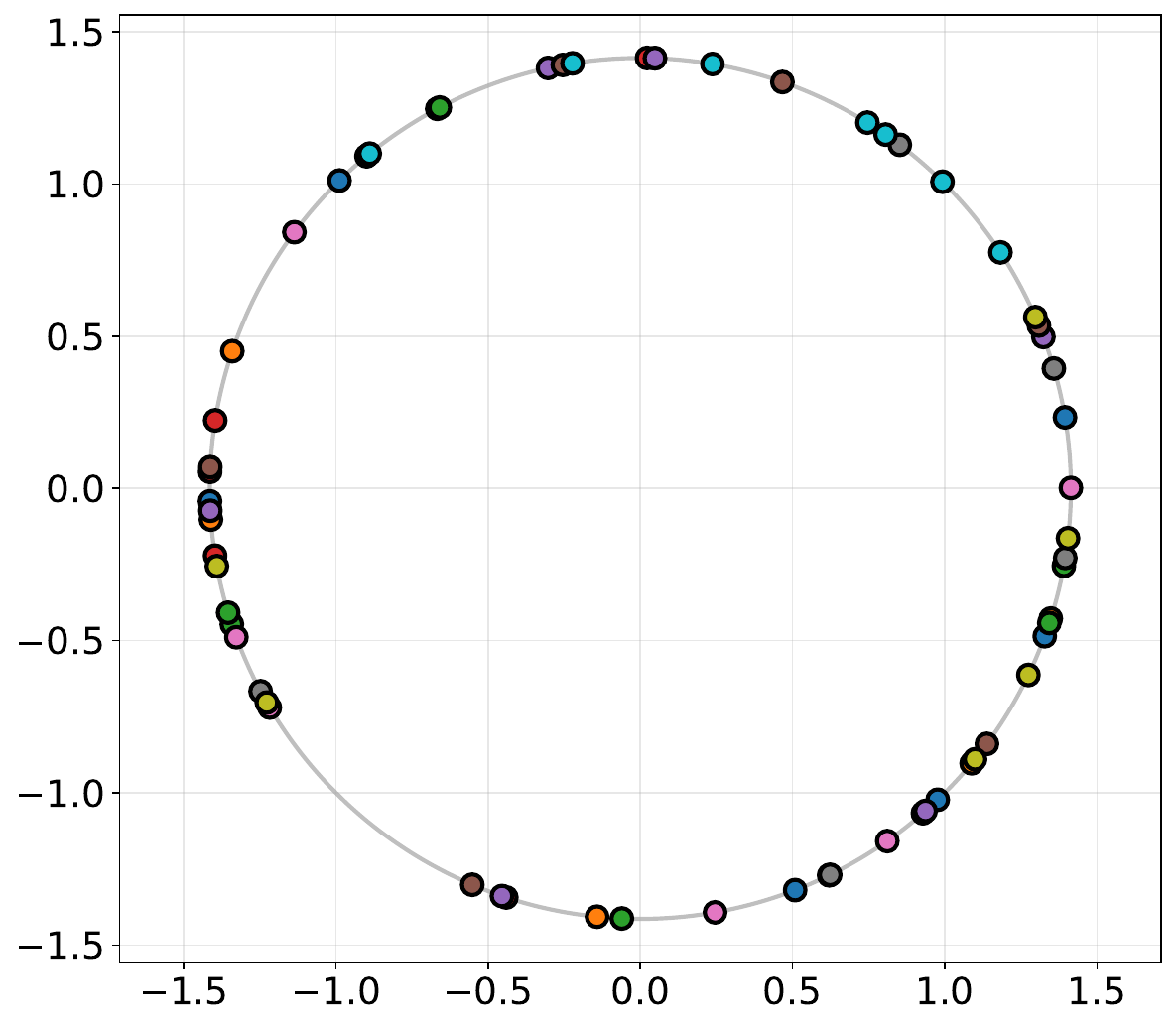}
        \caption{Initial positions.}
        \label{fig:simulation_initial}
    \end{subfigure}
    % \hfill
    \begin{subfigure}[t]{0.3\linewidth}
        \centering
        \includegraphics[width=\linewidth]{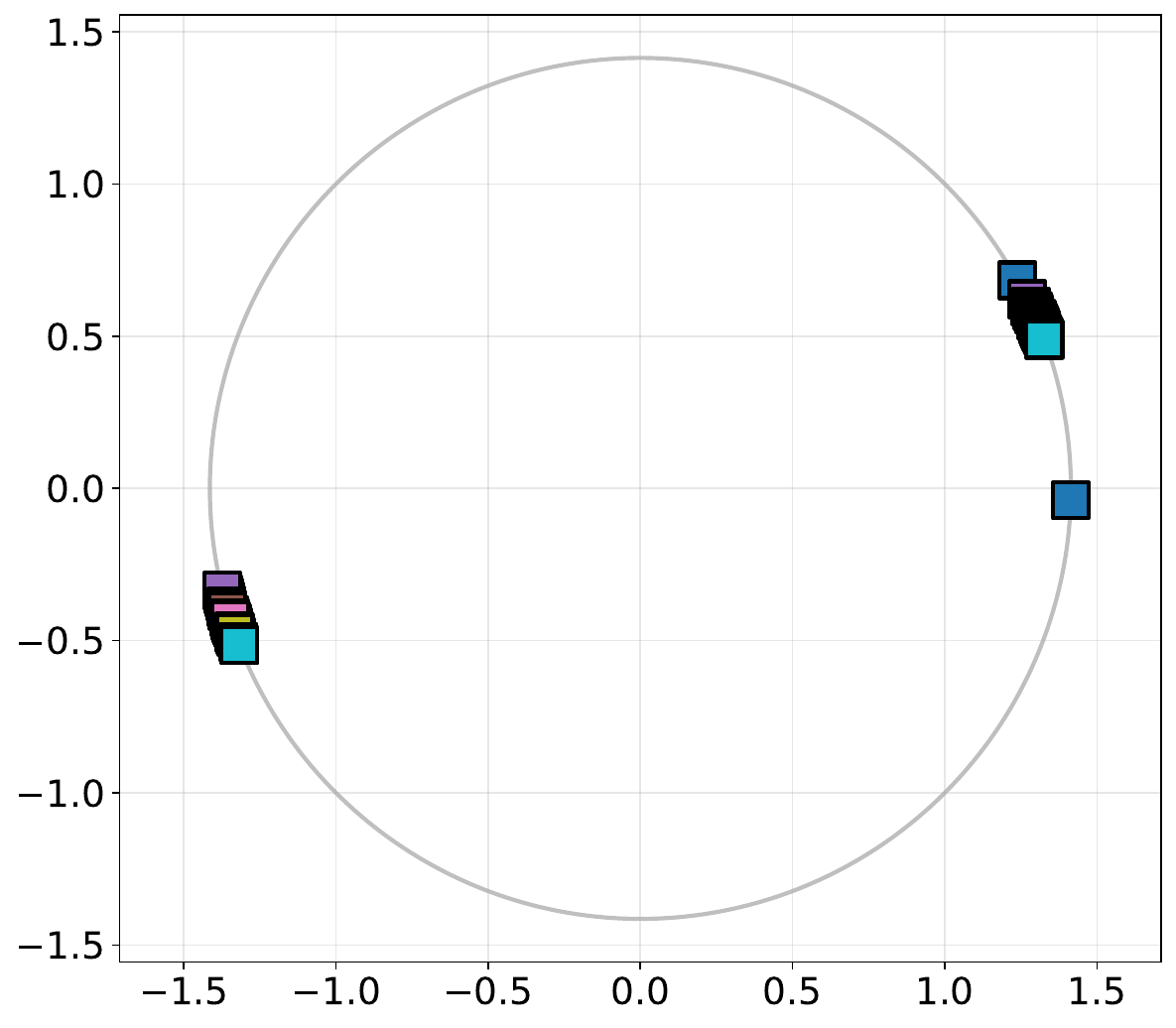}
        \caption{Final positions.}
        \label{fig:simulation_final}
    \end{subfigure}
    \caption{Initial (circles) and final (squares) token positions after processing through weight-sharing causal GQA with RMSNorm blocks. Tokens organize into two opposing clusters on the circle.}
    \label{fig:simulation}
\end{figure}

\begin{figure*}%[ht]
    \centering
    \begin{minipage}{0.19\linewidth}
        \centering
        \subfloat[Initial positions of tokens in 2D.]{\includegraphics[width=\linewidth]{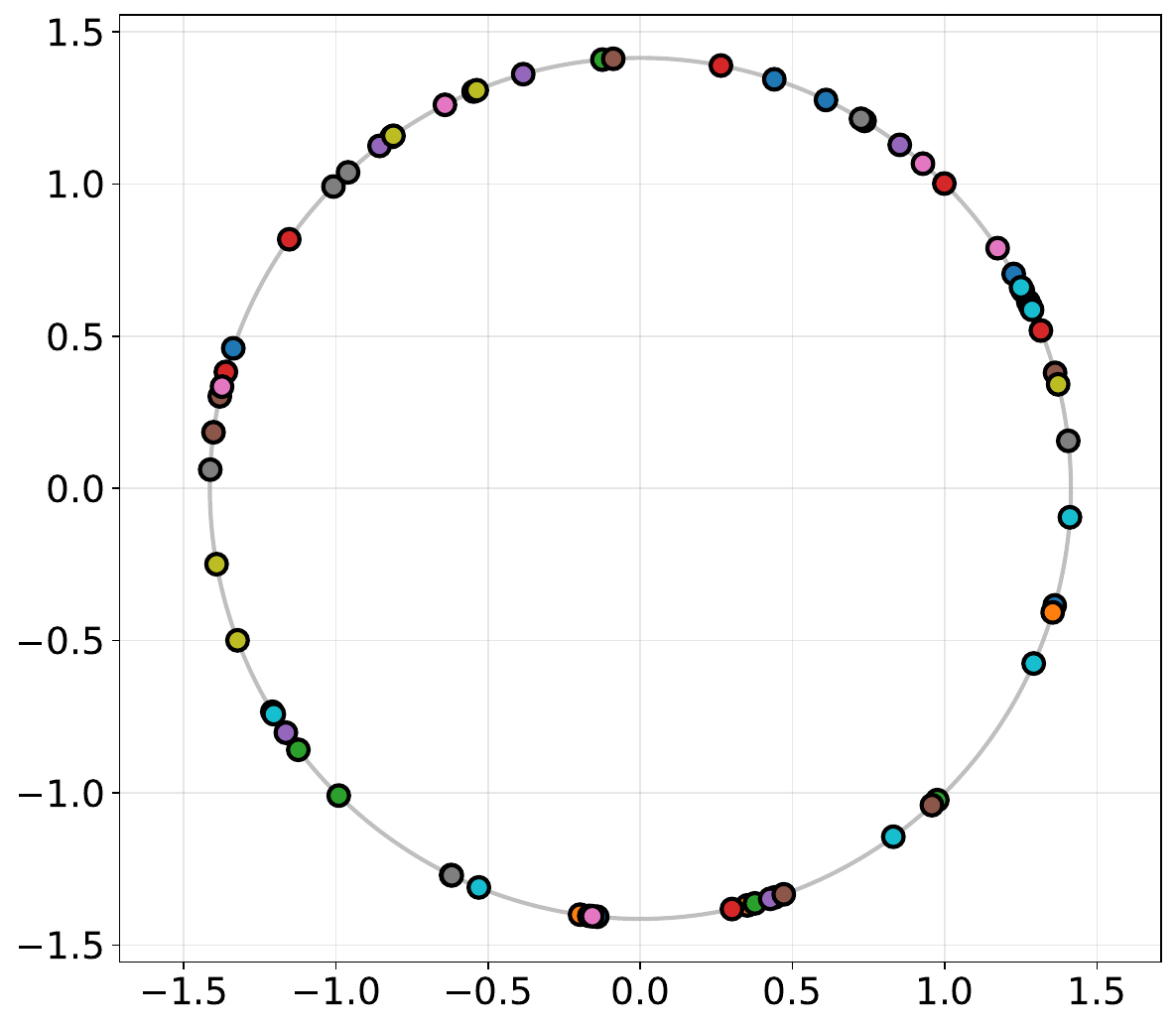}}
    \end{minipage}
    \hfill
    \begin{minipage}{0.19\linewidth}
        \centering
        \subfloat[Final positions of tokens in 2D.]{\includegraphics[width=\linewidth]{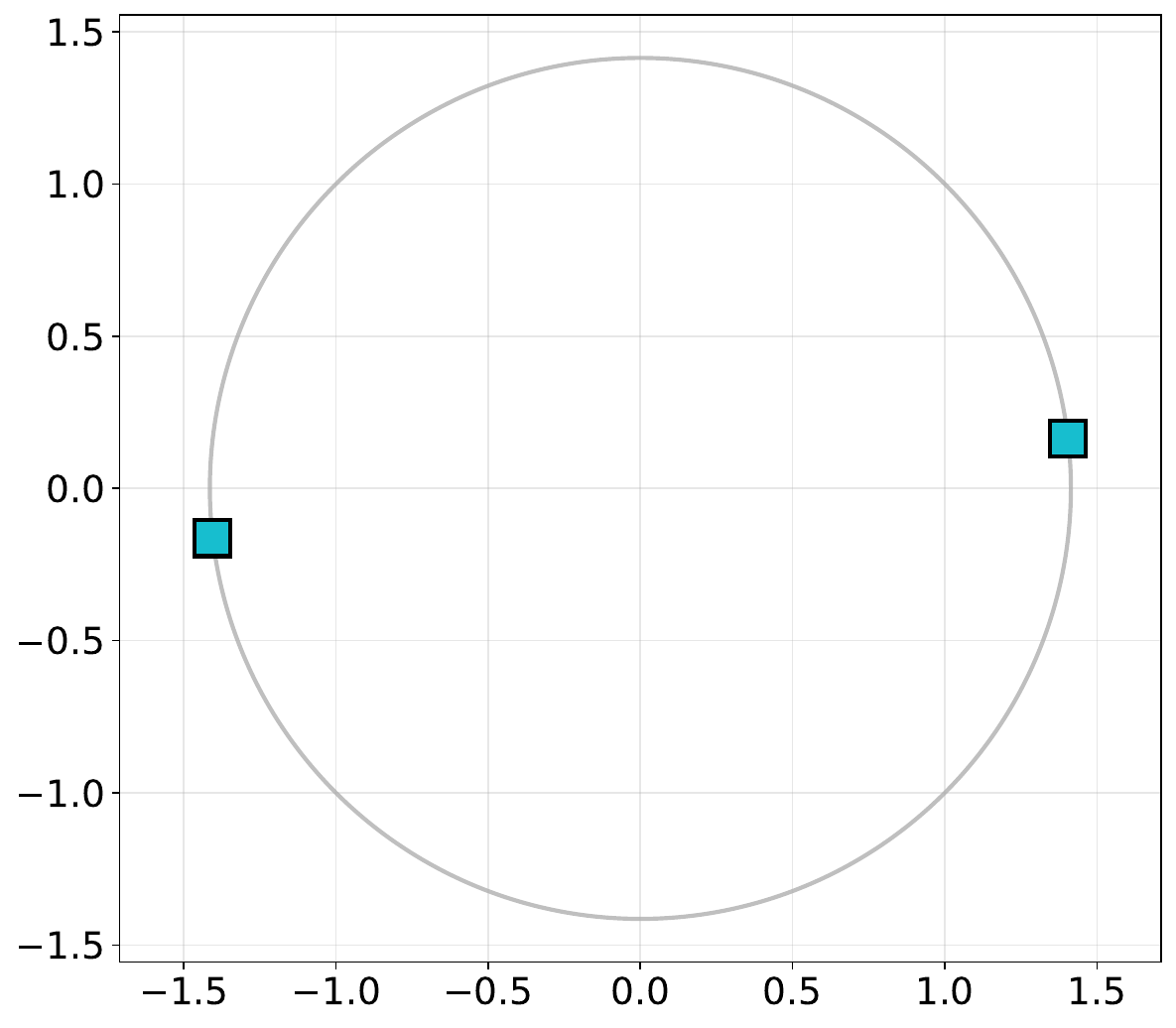}}
    \end{minipage}
    \hfill
    % \vskip\baselineskip
    \begin{minipage}{0.19\linewidth}
        \centering
        \subfloat[Token trajectories in 3D.]{\includegraphics[width=\linewidth]{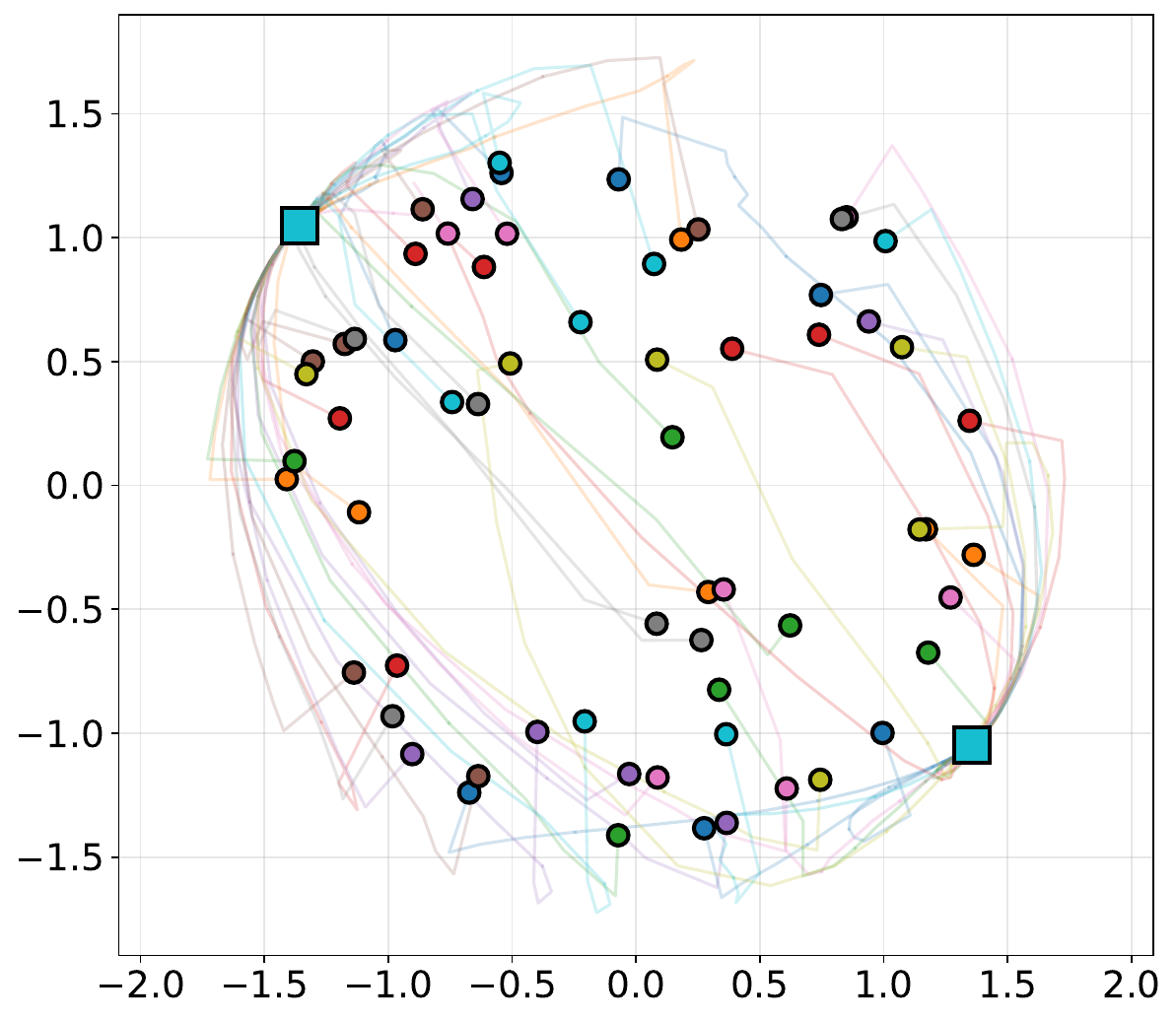}}
    \end{minipage}
    \hfill
    \begin{minipage}{0.19\linewidth}
        \centering
        \subfloat[Token trajectories in 32D.]{\includegraphics[width=\linewidth]{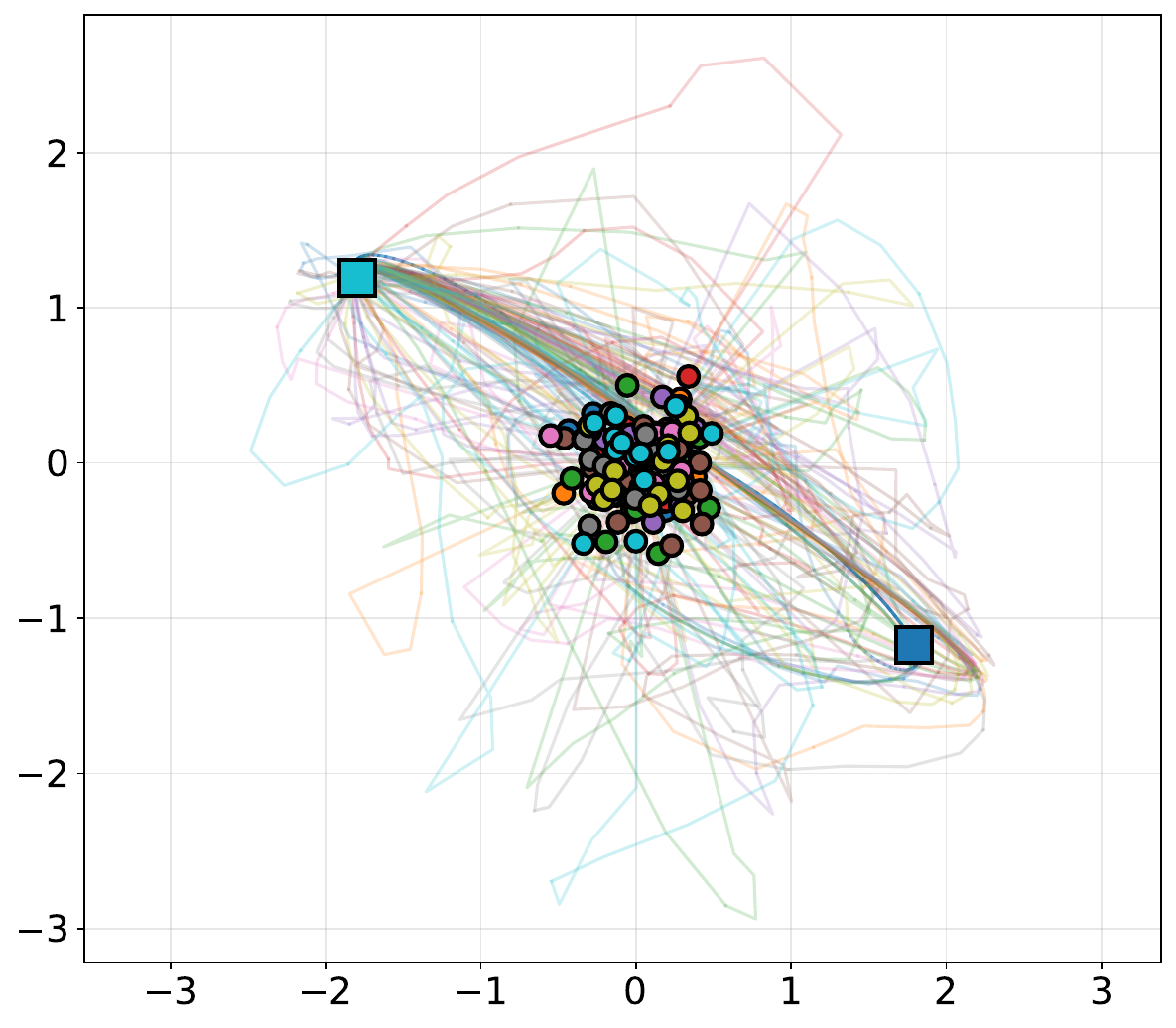}}
    \end{minipage}
    \hfill
    \begin{minipage}{0.19\linewidth}
        \centering
        \subfloat[Token trajectories in 128D.]{\includegraphics[width=\linewidth]{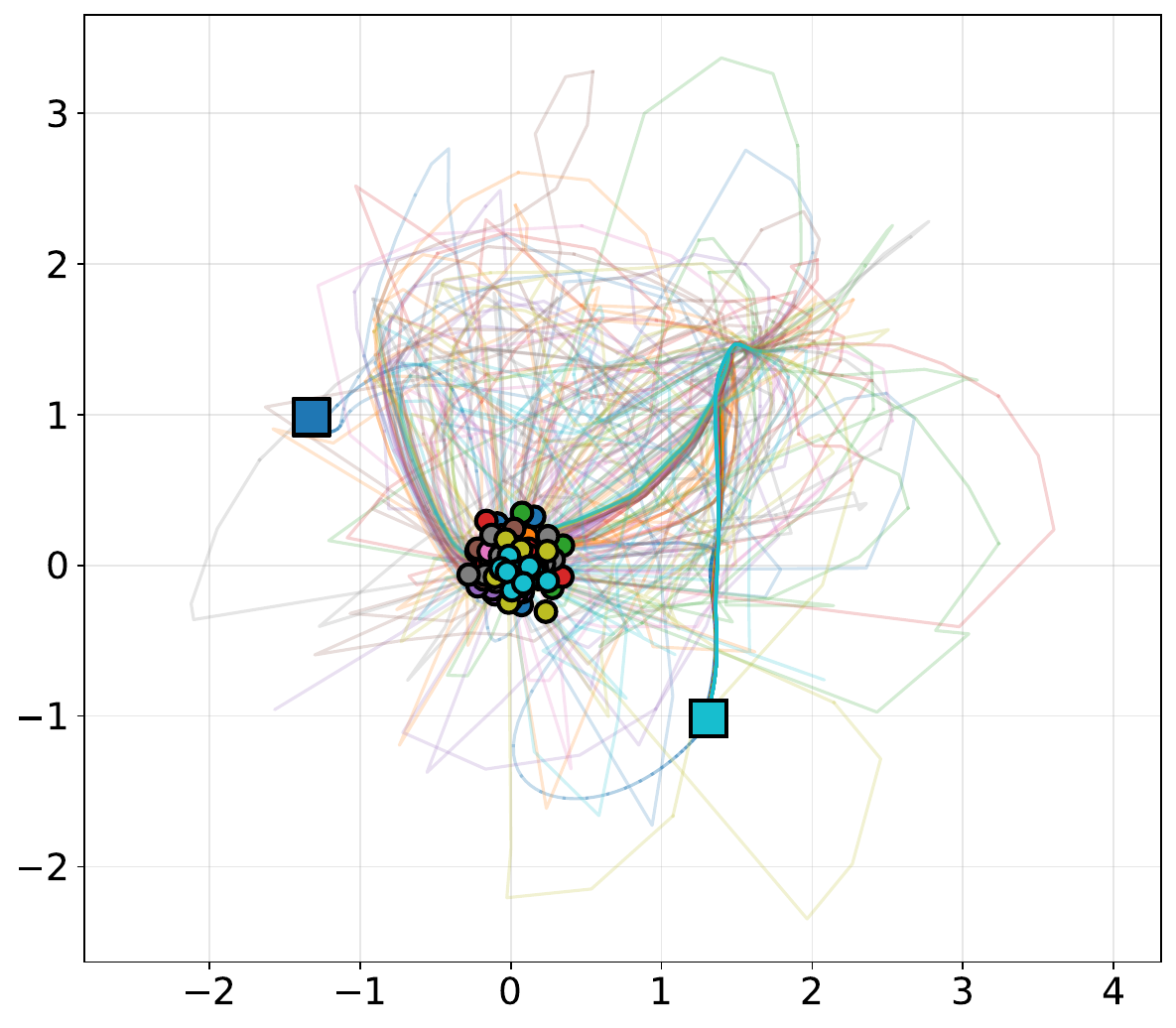}}
    \end{minipage}

    \caption{Tokens initial (circle) and final (square) positions and trajectories across various choices of hidden dimension.}
    \label{fig:token_trajectories_with_cond}
\vspace{-1em}
\end{figure*}

We begin by validating the condition for the two-cluster regime described in our paper. To this end, we conduct simulations using single-head attention with varying hidden dimensions, including 2D, 3D, 32D, and 128D. Each configuration employs 200 time-dependent layers that satisfy the assumptions outlined in Theorem~\ref{thm:two-dim-effective}. As shown in Figure~\ref{fig:token_trajectories_with_cond}, the results exhibit behavior consistent with the predicted behavior, that is, token representations converge toward one of two opposing points on the sphere.

To demonstrate that this phenomenon also arises in more sophisticated attention mechanisms, we then run a simulation using grouped-query attention (GQA)~\citep{ainslie2023gqa}, which shares key-value heads across groups of query heads to reduce KV-cache memory and accelerate inference. This attention mechanism is widely adopted in modern LLMs such as Llama3~\citep{dubey2024llama} and Mistral~\citep{jiang2023mistral7b}. We consider a model composed of 5000 weight-sharing blocks containing GQA and RMSNorm, operating in a 2-dimensional space, with 4 attention heads and a single KV head, each of dimension 2. The initial positions of 64 tokens in the sequence (Figure~\ref{fig:simulation_initial}) and all model parameters are randomly chosen. As illustrated in Figure~\ref{fig:simulation_final}, the final token positions tend to organize into two opposing clusters on the circle.

% \vspace{-0.2cm}
\subsection{Importance of the Value Projection Matrix}

Prior studies highlight the role of the value projection matrix $V$ in Transformer models, as it primarily affects information flow and training convergence~\citep{song2024unraveling,teo2024unveiling,boursier2025softmaxlinearattentionlargeprompt}. This property influences both LLM pre-training~\citep{zhou2025value, nguyen2023mitigating} and fine-tuning~\citep{hu2022lora, wang2025lorapro}. From our token-dynamic perspective in this work, we offer a novel explanation for this importance.

By Theorem~\ref{thm:two-dim-effective}, the convergence points \(\pm\xi_{\text{max}}\) of the token dynamics are the eigenvectors associated with the maximum eigenvalue of \(V\). These points emerge as stable attractors for every token in the sequence, dictating the long-term evolution of token representations. As a result, \(V\) plays a critical role in shaping token trajectories and thus significantly influences model behavior.

In line with this observation, ablation studies in \citep{hu2022lora} demonstrate that fine-tuning combinations including $V$ yields better performance. The experiment adapts GPT-3 175B to the WikiSQL~\citep{zhong2017wikisql} and MultiNLI~\citep{williams2018mnli} datasets, examining which subsets of self-attention weight matrices should be adapted. The number of trainable parameters is kept equal across configurations by inversely scaling the fine-tuning rank with the number of weights. We replicate these results in Table~\ref{tab:lora-attn-ablation} and observe that only fine-tuning $Q$ and/or $K$ fails to achieve optimal performance. This provides further evidence of $V$'s importance in fine-tuning strategies.

% LoRA~\citep{hu2022lora} and its variants~\citep{tay2022efficient, yen2025lora, wang2025lorapro, kopiczko2024vera, hao2024flora} have become widely used methods for adapting large language models (LLMs) to downstream tasks. LoRA works by training a low-rank differences matrices of the pre-trained weights of the model, allowing it to efficiently adapt to new tasks with fewer parameters compared to full fine-tuning methods. A key factor in their success lies in the value projection matrix \( V \), which plays a crucial role in token dynamics.

% \vspace{-0.4cm}
\begin{table}[t]
\centering
\caption{Validation accuracy (\(\%, \uparrow\)) on WikiSQL and MultiNLI when fine-tuning GPT-3 with LoRA applied to different subsets of self-attention weight matrices. All configurations use 18M trainable parameters. The standard deviation across random seeds is consistent for each dataset and reported in the first column. For the same number of fine-tuned weight types, combinations that include \(V\) yield higher accuracy. Results reproduced in \citep{hu2022lora}.}
\label{tab:lora-attn-ablation}
\begin{tabular}{lccccc}
\toprule
\cmidrule(lr){2-6}
Fine-tuning weight
& $Q$ & $K$ & $V$ & $Q, K$ & $Q, V$ \\
Fine-tuning rank
& 8 & 8 & 8 & 4 & 4 \\
\midrule
WikiSQL ($\pm 0.5\%$)
& 70.4 & 70.0 & 73.0 & 71.4 & 73.7 \\
MultiNLI ($\pm 0.1\%$)
& 91.0 & 90.8 & 91.0 & 91.3 & 91.3 \\
\bottomrule
\end{tabular}
\vspace{-1em}
\end{table}

% \section{Experiment}

% {\color{red} ... need experiments to verify the dimension reduction and two-clustering phenomena...}

%\vspace{-0.4cm}
\section{Conclusion}
%\vspace{-0.2cm}

In this paper, we introduce a new mathematical framework that partially resolves the two-cluster conjecture for causal attention with symmetric value matrices under arbitrary initial token configurations. In addition to establishing global convergence to $\pm \xi_{\max}$, our analysis provides the first rigorous explanation of the dimensional reduction phenomenon, which had previously been supported primarily by empirical observations. Central to our approach is a new analytical tool--which we term instability-within-the-flow--motivated by stability theory in physics. This perspective yields a principled route toward the conjecture, and it elucidates how geometric, dynamical, and statistical mechanisms jointly shape token dynamics. Overall, our results offer a theoretically grounded and practically useful characterization of clustering behavior in modern large-scale sequence models.

% \section*{Impact Statement}

% This paper presents work whose goal is to advance the field of machine learning. There are many potential societal consequences of our work, none of which we feel must be specifically highlighted here.

\begin{ack}
This research is supported by the National Research Foundation Singapore under the AI Singapore Programme (AISG Award No: AISG2-TC-2023-012-SGIL). This research is also supported by the Ministry of Education, Singapore, under the Academic Research Fund Tier 1 (FY2023) (A-8002040-00-00, A-8002039-00-00), the NUS Presidential Young Professorship Award (A-0009807-01-00), the NUS Artificial Intelligence Institute--Seed Funding (A-8003062-00-00), and the Cross Faculty Grant 2025, CFG25-012 (A-8004460-00-00). T. T. Nguyen acknowledges support from the NSFC Excellent Young Scientists Fund (Overseas) and the Research Start-up Fund of Shanghai Jiao Tong University.
\end{ack}

\bibliography{icml2026_conference}
\bibliographystyle{plain}

%%%%%%%%%%%%%%%%%%%%%%%%%%%%%%%%%%%%%%%%%%%%%%%%%%%%%%%%%%%%%%%%%%%%%%%%%%%%%%%
%%%%%%%%%%%%%%%%%%%%%%%%%%%%%%%%%%%%%%%%%%%%%%%%%%%%%%%%%%%%%%%%%%%%%%%%%%%%%%%
% APPENDIX
%%%%%%%%%%%%%%%%%%%%%%%%%%%%%%%%%%%%%%%%%%%%%%%%%%%%%%%%%%%%%%%%%%%%%%%%%%%%%%%
%%%%%%%%%%%%%%%%%%%%%%%%%%%%%%%%%%%%%%%%%%%%%%%%%%%%%%%%%%%%%%%%%%%%%%%%%%%%%%%
\newpage
\appendix
\onecolumn

\begin{center}
{\bf \Large{Appendix of ``Two-Clustering Regime of Token Dynamics in Causal Attention''}}
\end{center}

\DoToC

\section{Related Works}

\paragraph{ResNets and Neural ODEs}
Our work relates to the literature that views deep neural networks (DNNs) through a dynamical-systems perspective. In particular, standard DNNs can be interpreted as discrete-time dynamical processes in which representations propagate sequentially across layers \citep{lecun2015deep}. Residual networks (ResNets) \citep{chen2018neural, haber2017stable, kpotufe2012gradient} provide a canonical example of this viewpoint, while their continuous-time analogues--neural ordinary differential equations (neural ODEs) \citep{lin2018resnet,zhang2020approximation,xia2021heavy,li2022deep,nguyen2022improvingnode,thorpe2022grand,ruiz2023neural, cheng2023interpolation}--make the limiting transition to continuous time explicit. Under this formulation, the residual update $z = x + y(x,\theta)$ is understood as a discretization of a flow governed by an ordinary differential equation, which maps an initial condition $x(0)$ to a terminal state $x(T)$. A substantial body of work investigates the approximation, interpolation, and controllability properties of these architectures \citep{lin2018resnet, zhang2020approximation, li2022deep, tabuada2022universal, ruiz2023neural, cheng2023interpolation,teo2024momentumsmoe,nguyen2022improvingnode}.

%\vspace{-0.2cm}
\paragraph{Oversmoothing and Rank Collapse}
Transformers are susceptible to two prominent degeneracies: oversmoothing and rank collapse. Oversmoothing drives token representations toward indistinguishability, thereby reducing expressive capacity. In particular, Nguyen et al.~\cite{nguyen2023mitigating,nguyen2024coupled}, Wu et al.~\cite{wu2024demystifying}, Dovonon et al.~\cite{dovonon2024setting}, and Chen et al.~\cite{chen2022principle} show that self-attention alone is insufficient to avert this phenomenon. Proposed mitigations include \textit{ContraNorm}, a normalization scheme motivated by contrastive learning \citep{guo2023contranorm}, as well as residual connections coupled with normalization, which have been empirically shown to alleviate oversmoothing \citep{scholkemper2024residual}.

Rank collapse, by contrast, refers to the degeneration of hidden states onto a low-dimensional subspace. A growing body of work \citep{dong2021attention, feng2022rank, noci2022signal, joudaki2023impact, zhai2023stabilizing, noci2024shaped,nguyen2024pidformer, bao2024self} identifies this effect as a key bottleneck and develops stabilization mechanisms aimed at preserving representation diversity and improving downstream performance.

% \paragraph{Theoretical Analyses of Mamba}  
% Recent studies have begun to explore Mamba from a theoretical standpoint. Ali et al.~\cite{ali2024hidden} interpreted the S6 layer as a linear-time, softmax-free variant of attention, proving that Mamba is more expressive than transformers. In contrast, Muca Cirone et al.~\cite{cirone2025theoretical} applied Rough Path Theory to show that diagonal selective state space models (SSMs) are less expressive than non-diagonal ones. Merrill et al.~\cite{merrill2024illusion} used circuit complexity theory to argue that both SSMs and transformers belong to the class $TC^0$, implying computational equivalence. Jelassi et al.~\cite{jelassi2024repeat} provided both theoretical and empirical evidence that transformers surpass state space models in copying tasks. Together, these results form a theoretical basis for understanding the expressive and generalization limits of Mamba. 

%\vspace{-0.2cm}
\paragraph{Token Dynamics and Clustering Effects}
A complementary line of work models sequence architectures, including Transformers and Mamba, as interacting particle systems. Lu et al.~\cite{lu2019understanding} interpret self-attention as a numerical solver for a convection--diffusion equation, whereas Dutta et al.~\cite{dutta2021redesigning} introduce \textit{TransEvolve}, a particle-inspired design that reduces computational complexity. More broadly, this viewpoint places Transformer dynamics within the landscape of nonlinear collective-behavior models, such as flocking \citep{ha2019complete}, synchronization \citep{kuramoto1975self, acebron2005kuramoto}, and consensus dynamics \citep{krause2000discrete, motsch2014heterophilious}.

Within this dynamical framework, the asymptotic behavior of token trajectories has been investigated through clustering phenomena. Geshkovski et al.~\cite{geshkovski2024emergence} established convergence to limiting objects shaped by the initial conditions, and subsequent studies \citep{geshkovski2023mathematical, geshkovski2024dynamic} identified metastable multi-cluster configurations that can persist for extended times before ultimately merging into a single cluster. Alcalde et al.~\cite{alcalde2024clustering} further demonstrated asymptotic convergence to clustered equilibria for hardmax Transformers.  
Pham et al.~\cite{pham2025dynamical} investigated how different forms of positional encoding - specifically absolute and rotary - affect these dynamical regimes.
These results have been recently extended to causal attention in \cite{karagodin2024clustering}, all type of attentions in \cite{abellaconsensus}.
Furthermore, Vo et al.~\cite{vo2024demystifying} extended the analysis to the dynamical properties of tokens in a deep selective state-space model, and based on that, suggested improvements to the model by excluding negatively impactful scenarios and reordering tokens using importance scores.

% In contrast, the token dynamics of Mamba models remain largely unexplored. 
% Existing analyses impose restrictive assumptions - such as one-dimensional embeddings or the absence of normalization layers - that oversimplify practical implementations~\cite{vo2024demystifying}. 

% In this paper, we take one step further by studying the token dynamics for Mamba without these restrictive assumptions.

% \section{Attention Dynamic - A Two-Clustering Phenomena}

\section{Setting}

In this paper, we study the dynamical properties of the system
\beq\label{main-eq-now}
\pt_t x_i=\sum_{j=1}^i a_{ij}(t)\,\P_{x_i}(Vx_j),
\eeq
where $x_i:[0,\infty)\to \S^{d-1}$ denotes the $i$-th token evolving on the unit sphere $\S^{d-1}\subset\R^d$, with $d\ge 2$. The projection operator $\P$ is defined by
\[
\P_{y}(x)=x-(x\cdot y)y,\qquad y\in \S^{d-1}.
\]
The matrix $V\in\R^{d\times d}$ is referred to as the \emph{value matrix}. For notational simplicity, we assume throughout that
\[
V(t)=\diag\{\lambda_1(t),\lambda_2(t),\dots,\lambda_d(t)\}, 
\]
so that the eigen-basis coincides with the canonical basis $\{\vec e_1,\dots,\vec e_d\}$ and 
\beq\label{order-V}
\lambda_1(t)>\lambda_2(t)>\cdots>\lambda_d(t)\quad\text{for all}\quad t\ge 0.
\eeq
We denote by $Q(t),K(t)\in\R^{d\times d}$ the \emph{query} and \emph{key} matrices, and define
\[
Z_i(t)=\sum_{j=1}^i e^{\beta \la Qx_i(t),Kx_j(t)\ra},
\]
where $\beta>0$ is a fixed parameter. The coefficients $a_{ij}$ are then given by
\[
a_{ij}(t)=a_{ij}(x_i(t),x_j(t))=\frac{e^{\beta \la Qx_i,Kx_j\ra}}{Z_i(t)}.
\]
Define $Q(t)\tp K(t)=\mathcal T(t)$, so that
\[
a_{ij}(t)=\frac{e^{\beta \la x_i, \mathcal Tx_j\ra}}{Z_i(t)},\qquad 
Z_i(t)=\sum_{j=1}^i e^{\beta \la x_i,\mathcal Tx_j\ra}.
\]
In particular, the evolution equation for the $L$-th token reads
\[
\pt_t x_L=\sum_{j=1}^L a_{Lj}\,\P_{x_L}(Vx_j).
\]
The dynamical systems perspective---in which tokens are viewed as interacting particles---has attracted considerable interest from both the machine learning and mathematical communities. This framework provides a mathematically rigorous lens for understanding how Transformers internally organize information through token interactions. The synthesis of geometric, dynamical, and statistical insights makes the characterization of token dynamics both theoretically elegant and crucial for interpreting modern large-scale models.

\paragraph{Token Dynamics for Causal Attention.}
Token dynamics for causal attention have been studied recently in \cite{karagodin2024clustering} for the time-independent setting and in \cite{abellaconsensus} for the time-dependent setting. Since both works are closely related to our analysis, we briefly review their main results below.

\begin{itemize}
    \item \textit{Time-Independent \& Causal Attention:} Karagodin et al.~\cite{karagodin2024clustering} established the convergence of the first token $x_1(t)$ for arbitrary constant value matrices $V$ and proved global convergence of all tokens to $x_1(0)$ in the special case $V = I_d$. Their analysis leverages explicit closed-form solutions and tools from generic dynamical systems (Schub genericity theorem). Importantly, they also predicted and conjectured a dimensional deduction mechanism based on experimental observations of real Transformer value matrices. 
    The work of Karagodin et al.~\cite{karagodin2024clustering} further conjectured this dimensionality reduction mechanism and highlighted its practical implications:
    \begin{quote}
    \itshape
    An important practical consequence of these conjectures is that they describe a dimensionality reduction mechanism induced by the value matrix $V$. Token trajectories converge to $L \cap \mathbb S^{d-1}$ and, once restricted to this set, evolve as if governed by the scalar matrix $\lambda I_d$ on the smaller sphere $\mathbb S^{\dim L - 1}$. For the pre-trained model of Lan et al.\ (2020), the spectra of the value matrices are shown in Figure~6, where some heads even exhibit a negative top eigenvalue $\lambda_{\max}$. A natural direction for future work is to analyze value matrices $V$ arising in real-world networks and relate their leading eigenspaces to the semantic roles of layers and tokens.
    \end{quote}

    \item \textit{Fully Time-Dependent Case:} Abella et al.~\cite{abellaconsensus} analyzed the causal model with fully time-dependent matrices $V(t), K(t), Q(t)$. Assuming all initial tokens lie within a single hemisphere of $\mathbb{S}^{d-1}$, they employed a Lyapunov argument to prove global convergence to a single cluster $\xi_{\max}$, fully resolving the single-cluster regime. The hemisphere condition is preserved under the dynamics, enabling a clean analytical framework, but restricts admissible initial configurations rarely satisfied in practice.
    However, their results do not resolve the conjecture established in \cite{karagodin2024clustering}.
\end{itemize}

In contrast, our work offers a mathematically rigorous lens for explaining how Transformers internally organize information through token interactions. 
To this end, we introduce a new analytical tool--which we term instability-within-the-flow--inspired by stability theory in physics. 
This framework enables us to partially resolve the conjecture and clarifies the synthesis of geometric, dynamical, and statistical perspectives underlying token dynamics. 
As a result, our characterization is both theoretically principled and practically informative for interpreting modern large-scale models.

\paragraph{Structure of the proofs and key ideas .} Our first step is to derive the leading-order dynamics of the \(L^{\text{th}}\) token, assuming that all preceding tokens have converged to \(\pm \xi_{\max}\). This analysis is carried out in Section~\ref{leading-sec} and substantially simplifies the dynamics of the final token. Next, for notational convenience, we assume \(\xi_{\max} = \vec e_1\), which can be achieved without loss of generality by applying a unitary rotation to the solution. We then explicitly derive the coordinate-wise evolution equations for the \(L^{\text{th}}\) token, as stated in Lemma~\ref{coor-wise}. These steps establish a natural and systematic setup for studying the dynamics of the final token. 

Our first key result is presented in Theorem~\ref{sup-main}. Roughly speaking, we uncover a high-dimensional reduction mechanism in which the dynamics collapse to \emph{exactly two coordinates} for arbitrary matrices \(Q\) and \(K\).
Our key idea is to \emph{separate the direction dynamics from the magnitude dynamics}: once we factor out the overall magnitude of a block of coordinates, the remaining dynamics describe a competition over which direction the token points to.
Because different coordinates have systematically different growth advantages (encoded by the ordered of the eigenvalues \eqref{order-V} of the value matrix $V$), this directional dynamics has a built-in drift that continuously sharpens the representation, pushing it toward the boundary where some coordinates vanish.
As a result, the system cannot remain in a persistent mixed state: either a given coordinate loses mass and is suppressed, or it survives and eventually dominates, eliminating all weaker competitors.
The only subtlety is noise inherited from earlier tokens, but as long as this residual signal decays faster than the scale of the current competition, it cannot prevent concentration.
In effect, the dynamics behave like an expert-selection process: coordinates are tested one by one, and any direction that remains consistently active must ultimately win decisively.
This explains why, despite the ambient dimension, the token collapses onto a low-dimensional structure driven by a small number of dominant features. The full detail of the proof of Theorem \ref{sup-main} is presented in Section \ref{reduce-2d}.

Based on Theorem~\ref{sup-main}, we develop our theory in two separate directions: 
\begin{itemize}

\item \underline{Structured time-dependent $Q(t),K(t)$ (Theorem~\ref{time-dependent-main}).}
Under suitable block-structure assumptions on $Q(t)$ and $K(t)$, we derive an effective ODE governing the evolution of the projection $x_L \cdot \xi_{\max}$ (later denoted by $y_1$), treating all remaining coordinates as perturbations. This reduction is carried out in Theorem~\ref{all-zero}. Formally, the dynamics of $y_1$ take the form
\begin{equation}\label{y1-dyn}
\partial_t y_1 = (1 - y_1^2)\, M(t,y_1) + o(1),
\end{equation}
where $o(1)$ denotes a quantity that vanishes as $t \to \infty$. Since the equation is non-autonomous and the right-hand side is not monotone for the  two-cluster regime $\{\xi_{\max}, -\xi_{\max}\}$, we cannot linearize around a stationary solution and apply the stable manifold theorem as in \cite{karagodin2024clustering,geshkovski2023mathematical}, nor use the Lyapunov method as in \cite{alcalde2024clustering}. Instead, the right-hand side of \eqref{y1-dyn} admits a  ``moving equilibrium'' in the interior of $(-1,1)$—a phenomenon already visible in the simpler setting \eqref{toy-1}. This behavior corresponds precisely to the moving-hilltop analogy in the stability theory of physical systems, which we discuss in Section~\ref{instability}. Our key insight is that, under suitable time-varying assumptions on the parameter matrices, this unstable equilibrium cannot trap the solution $y_1$ in the interior of $(-1,1)$. Consequently, $x_L$ must converge to one of the two clusters: $\xi_{\max}$ (corresponding to $y_1 = 1$) or $-\xi_{\max}$ (corresponding to $y_1 = -1$). The full analytical framework is developed in Theorem~\ref{almost} and Section~\ref{full-proof}. Importantly, our analysis is not limited to establishing convergence to the two-cluster configuration. Rather, it also provides insight into alternative clustering behaviors that may arise in practice and elucidates the instability mechanisms that can unlock such dynamics. In particular, we present an example in which oscillatory drifting of the unstable equilibrium within the interior can trap the $L$-th token and lead to dynamics distinct from convergence to $\pm \xi_{\max}$. This demonstrates that our assumptions are essentially sharp; see Proposition~\ref{trap-case}.

\item \underline{General stationary $Q,K$ (Theorem~\ref{generalQK}).}
Once the parameter matrices $Q$ and $K$ have stabilized, the dimension-reduction result of Theorem~\ref{sup-main1} shows that, asymptotically, the token dynamics are confined to the manifold
\[
\mathbb{S}^1 \times \{0\}^{d-2}.
\]
This residual degree of freedom corresponds to a phase (or orientation) ambiguity in the embedding space. A natural question is whether the token may continue to rotate along this manifold indefinitely, even after the training dynamics appear to have converged. Using tools from complex analysis, we prove that the set of admissible long-time limit points on $\mathbb{S}^1$ is necessarily discrete. Consequently, the token representation must converge to a single fixed configuration (see Lemma~\ref{son-lem-3}). In particular, apparent stabilization of the parameter matrices $Q$ and $K$ implies convergence of the token dynamics. We refer to Theorem~\ref{generalQK} and Section~\ref{proofQK} for the complete argument.

\end{itemize}

\section{Main Results}

\begin{theorem}\label{time-dependent-main}
Assume that $\mathcal T(t)=Q^\top K(t)$ has the block structure
\begin{equation}\label{T-condition}
\mathcal T(t)=
\begin{pmatrix}
m_0(t) & 0 \\
0 & m_1(t) I_{d-1} + \mathcal B(t)
\end{pmatrix},
\qquad 
\mathcal B(t)^\top=-\mathcal B(t),
\end{equation}
for some functions $m_0(t)>0$ and
\begin{equation}\label{assumption1}
m_1(t)<m_0(t)+\frac{1}{2\beta}
\qquad\text{for all }t\ge0.
\end{equation}
Assume further that
\[
\lim_{t\to\infty}(V(t),\mathcal T(t))=(V_\infty,\mathcal T_\infty),
\qquad
\lim_{t\to\infty}V'(t)=0,
\]
where
\[
V_\infty=\diag\{\lambda_{1,\infty},\ldots,\lambda_{d,\infty}\},
\qquad
\lambda_{1,\infty}>\lambda_{2,\infty}>\cdots>\lambda_{d,\infty},
\qquad
\lambda_{1,\infty}>0.
\]
Suppose that
\[
x_i(t)\longrightarrow \pm \vec e_1
\qquad\text{as }t\to\infty,
\quad i=1,\ldots,L-1.
\]
Then either of the following must hold
\begin{itemize}
    \item There exists a sequence $t_n\to \infty$ such that $x_L(t_n)\to \pm \vec e_1$
    \item Then there exists an index $\ell\in\{2,\ldots,d\}$ such that the following holds.
For every solution $x_L(t)$, all transverse components except one decay,
\begin{equation}\label{transverse-decay}
x_L(t)\cdot\vec e_j \longrightarrow 0
\qquad\text{as }t\to\infty,
\quad \forall\, j\neq 1,\ell,
\end{equation}
so that the long--time dynamics collapses onto the two--dimensional subspace
$\mathrm{span}\{\vec e_1,\vec e_\ell\}$.

Within this reduced dynamics, a conditional uniqueness property holds:
if $x_L^{(1)}(t)$ and $x_L^{(2)}(t)$ are two solutions (possibly with different
initial data) whose transverse components coincide,
\begin{equation}\label{same-transverse}
x_L^{(1)}(t)\cdot \vec e_j \equiv x_L^{(2)}(t)\cdot \vec e_j
\qquad \forall\, t\ge0,\ \forall\, j\neq 1,\ell,
\end{equation}
then, except possibly for a single exceptional trajectory in this constraint
class, the corresponding solution satisfies
\[
x_L(t_n)\longrightarrow \pm \vec e_1
\qquad\text{along some sequence }t_n\to\infty.
\]
In particular, apart from this exceptional orbit, all solutions
\emph{recurrently return} to the dominant direction in the infinite--time limit.
\end{itemize}
\end{theorem}

\begin{remark}
We show that, after stabilization of the earlier tokens, the dynamics of the
remaining token $x_L(t)$ admits only two long--time possibilities. Either the
trajectory visits the dominant direction $\vec e_1$ infinitely many times as
$t\to\infty$, or the dynamics collapses onto a two--dimensional invariant
subspace. Within this reduced two--dimensional regime, and under a natural
conditional uniqueness assumption, all trajectories except possibly a single
exceptional one recurrently return to $\vec e_1$. In this sense, our theorem
provides a principled theoretical explanation for the emergence of $\vec e_1$
as the dominant direction, even in the presence of time--varying attention.
\end{remark}

\begin{remark}[On the structural assumption for $\mathcal T(t)=Q^\top K(t)$]
The block structure in \eqref{T-condition} ensures that the distinguished direction $\vec e_1$, corresponding to the
leading eigenvalue of $V_\infty$, remains an invariant axis under the
$Q^\top K(t)$–induced interaction.  No mixing occurs between $\vec e_1$ and the
transverse coordinates; all nontrivial coupling is confined to the orthogonal
$(d-1)$–dimensional subspace through the skew–symmetric component
$\mathcal B(t)$.  
The coefficients $m_0(t)$ and $m_1(t)$ control the relative strength of the
interaction along and perpendicular to $\vec e_1$, while $\mathcal B(t)$
contributes only rotational effects within the transverse space.
This decomposition isolates the dominant direction dynamically and is precisely
the structure needed for the reduction to a one–dimensional effective equation
along $y_1$.
\end{remark}

\begin{remark}
The restriction \eqref{same-transverse} is not artificial.
By part~(i), all transverse coordinates except $\vec e_\ell$ decay asymptotically and
are therefore slaved to the dynamics in $\mathrm{span}\{\vec e_1,\vec e_\ell\}$.
The conditional uniqueness statement in part~(ii) should thus be understood as a
fiber--wise result: for each fixed transverse profile, the reduced two--dimensional
dynamics admits at most one non--dominant trajectory.
\end{remark}

\begin{remark}
The structural assumptions
\[
\lambda_1(t)> 0, 
\qquad 
m_0(t)>0, 
\qquad 
m_1(t)<m_0(t)+\frac1{2\beta}
\]
can in fact be replaced by the more general asymptotic condition
\[
\lambda_{1,\infty}\, \beta m_{0,\infty} L
\;+\;
(\lambda_{1,\infty}-\lambda_{k,\infty})
\min\Bigl\{1,\;1-2\beta(m_{1,\infty}-m_{0,\infty})\Bigr\}
\;>\; 0,
\qquad\text{for all } k\ge 2.
\]

However, to keep the presentation streamlined and the exposition focused, 
we will restrict attention to the simpler set of hypotheses stated in the theorem, 
under which all essential ideas already appear transparently.
\end{remark}

\begin{theorem}[Hierarchical Suppression Mechanism]\label{sup-main}
Let $y(t)\in\mathbb S^{d-1}$ be the $L$-th token, and let $\varepsilon_{L-1}(t)$ denote 
the residual error contributed by the first $L-1$ tokens.  
For each $\ell\in\{2,\dots,d\}$, define
\[
R_\ell(t)=\Bigl(\sum_{j=\ell}^d y_j(t)^2\Bigr)^{1/2},
\qquad
\kappa_\ell(t)=\frac{\varepsilon_{L-1}(t)}{R_\ell(t)}.
\]
Suppose that for some $\ell\ge2$,
\begin{equation}\label{con-1}
    \kappa_\ell(t)\longrightarrow 0,
    \qquad (t\to\infty).
\end{equation}
Then exactly one of the following two alternatives occurs:

\medskip
\noindent\emph{(A) $y_\ell$ becomes the dominant transverse direction.}  
\[
y_j(t)\longrightarrow 0,
\qquad\text{for all } j\neq 1,\ell.
\]
In this case the suppression process terminates at level $\ell$.

\medskip
\noindent\emph{(B) $y_\ell$ is suppressed.}
\[
y_\ell(t)\longrightarrow 0.
\]
Here the transverse vector $(y_{\ell+1},\dots,y_d)$ remains nontrivial, and the 
same mechanism applies at the next level $\ell+1$ using 
$\kappa_{\ell+1}(t)$.

\medskip

Thus, whenever the error-to-transverse ratio becomes negligible at level $\ell$, the 
dynamics eliminate all but one of the coordinates $y_\ell,\dots,y_d$.  
Iterating this reduction yields a unique surviving transverse coordinate $y_k$, and the 
trajectory collapses onto the $(\ell)$-dimensional subsystem spanned by
\[
(y_1, y_2, \dots, y_{\ell-1}, y_k).
\]
\end{theorem}

\begin{theorem}\label{generalQK}
Let $Q$ and $K$ be arbitrary matrices.  
Assume that $x_i(t)\to \pm \vec e_1$ for all $i\le L-1$, and that the DDSC 
condition~\eqref{DDSC} holds for the final $L^{th}$ token.  
Then there exists a finite subset $\mathcal{S}\subset\mathbb{S}^{d-1}$ such that,
for almost every initial condition, the trajectory $x_L(t)$ converges to some 
$y_\star\in\mathcal{S}$ as $t\to\infty$.

Moreover, if the limit $y_\star$ is not equal to $\pm\vec e_1$, then $y_\star$ 
must lie in a two–dimensional coordinate plane and take the form
\[
(\alpha_1,0,\dots,0,\al_\ell,0,\dots,0),
\qquad \alpha_1^2+\al_\ell^2=1,
\]
that is, only the first and the $\ell$-th coordinates are nonzero.  
In this case the limit point $y_\star$ satisfies the scalar algebraic condition
\[
G(y_\star)=0,
\qquad \text{for some } \ell\in\{2,\ldots,d\},
\]
where
\[
G(y)
= e^{\beta\, y\cdot \mathcal{T}y}\,(\lambda_1-\lambda_\ell)y_1
 \;+\;\lambda_1\!\left[N_1 e^{\beta\, y\cdot\mathcal{T}\vec e_1}
     - N_2 e^{-\beta\, y\cdot\mathcal{T}\vec e_1}\right].
\]
\end{theorem}

\begin{remark}
Theorem~\ref{generalQK} reveals an unexpectedly rigid structure in the 
high-dimensional token dynamics.  
Although the system is nonlinear and a priori allows infinitely many potential 
limit configurations, the theorem shows that the long-time behavior is governed 
by a \emph{finite} set of isolated equilibria.  
This finiteness holds for \emph{all} choices of $Q$ and $K$ and for 
\emph{almost every} initial condition. Even more striking is that every nontrivial limit point $y_\star$ is forced  to lie on a two-dimensional coordinate plane and is completely characterized by  the scalar equation $G(y_\star)=0$.  
Thus, a high-dimensional nonlinear problem collapses to solving a single 
one-dimensional equation and checking the sign of the eigenvalues of a 
$2\times 2$ linearized system.  
\end{remark}

\begin{theorem}[Sharpness of the condition $\lambda_1> 0$] \label{toy-case}
Suppose $\lambda_1<0$. Then there exist $\beta>0$ and $\lambda_2<\lambda_1$ such that, for $L=d=2$, $\mathcal T=I_2$, and $x_1(t)\equiv \vec e_1$, one can find a nonempty open set $U\subset \S^{1}$ with the property that for all $x_2(0)\in U$, the limit 
\[
\lim_{t\to\infty} x_2(t)
\]
exists and satisfies 
\[
\lim_{t\to\infty} x_2(t)\notin \{\pm \vec e_1\}.
\]
\end{theorem}
\begin{proof}
We write 
$
x_2(t)=x(t)\vec e_1+y(t)\vec e_2,
$. From the equation $
\pt_t x_2=a\,\P_{x_2}(Vx_2)+b\lambda_1 \P_{x_2}(\vec e_1),
$ and the fact that $x_2\in\S^1$ for all time, we deduce 
\[
\begin{aligned}
x'(t) 
&= (1-x^2)\,\frac{g(x)}{e^\beta+e^{\beta x}},\quad g(x)=e^\beta(\lambda_1-\lambda_2)x+\lambda_1 e^{\beta x}.
\end{aligned}
\]
Thus, the first coordinate of $x_2$ satisfies a scalar ODE taking values in $(-1,1)$.  
In particular, if $g$ admits two distinct roots $r_1,r_2\in(-1,1)$, then by the vector field method we obtain
\[
\lim_{t\to\infty}x(t)\in \{r_1,r_2\}
\]
whenever the trajectory $x(t)$ remains in the interval $(r_1,r_2)$. We now show that such roots exist when $\lambda_1<0$ and $\beta$ is chosen appropriately.  
Suppose $r\in(-1,1)$ is a root of $g$. Since $\lambda_1<0$ and $\lambda_1-\lambda_2>0$, it follows that $r>0$.  
The root condition $g(r)=0$ is equivalent to
\begin{equation}\label{ran-23}
c(\beta r)=\gamma,
\qquad c(x)=xe^{-x}, 
\qquad \gamma=\frac{|\lambda_1|\beta}{e^\beta(\lambda_1-\lambda_2)}=\frac{|\lambda_1|\beta}{e^\beta(|\lambda_2|-|\lambda_1|)}.
\end{equation}
Since $c'(x)=e^{-x}(1-x)$, the function $c$ is strictly increasing on $(0,1)$ and strictly decreasing on $(1,\infty)$.  We will show that when $\beta>1$ and $\gamma\in(\beta e^{-\beta},e^{-1})$, the equation $c(x)=\gamma$ has two solutions, one solution in $(0,\beta^{-1})$ and one solution in $(\beta^{-1},1)$.   ~\\ 
As $r\in(0,1)$, we examine solutions of $c(\bar r)=\gamma$ with $\bar r\in(0,\beta)$.  ~\\
Indeed, because $\gamma\in(c(0),c(1))=(0,e^{-1})$ and $c$ is increasing on $(0,1)$, there exists a solution $\bar r_1\in(0,1)$.   
On $(1,\beta)$, since $c$ is decreasing and $\gamma\in(c(\beta),c(1))=(\beta e^{-\beta},e^{-1})$, there exists another solution $\bar r_2\in(1,\beta)$.  ~\\
Consequently,
\[
g(r_1)=g(r_2)=0,
\qquad r_1=\tfrac{\bar r_1}{\beta}\in(0,\beta^{-1}),\qquad 
r_2=\tfrac{\bar r_2}{\beta}\in(\beta^{-1},1).
\]
Now if we take any data $x(0)=x_2\cdot\vec e_1\in (r_1,r_2)$, we have $x(t)\to x_\star\in \{r_1,r_2\}$. 
This completes the proof. 
\end{proof}

\section{Deriving the Leading Dynamics}\label{leading-sec}
We define the error function
\[
\eps_{L-1}(t)=\sum_{j=1}^{L-1} \bigl\|x_j(t) - s_j \vec e_1 \bigr\| 
\]
 Our inductive hypothesis for the first $L-1$ tokens is
\begin{equation}\label{induction}
\eps_{L-1}(t)\;\longrightarrow\; 0
\qquad \text{as } t \to \infty,
\end{equation}
Here $s_j$ simply records \emph{which cluster} the token $x_j$ converges to: $+1$ 
corresponds to the ``top'' copy of $\vec e_1$ and $-1$ to the ``bottom'' copy 
$-\vec e_1$. 
To simplify the notation, in this paper we denote 
\[
y=x_L
\]
to be the $L^{th}$ token.
Our goal is to identify the leading dynamics of $y$. The main results are stated in Theorem \ref{sup-main}. We assume \eqref{induction}. 
The evolution equation for $y=x_L$ reads
\begin{equation}\label{y-eq}
\pt_t y \;=\; a_{LL}\P_y(Vy)+\sum_{j=1}^{L-1} a_{Lj}\P_y(Vx_j).
\end{equation}
Here
\[
a_{LL}=\frac{e^{\beta \la y,\mathcal Ty\ra}}{Z_L(t)}, 
\qquad 
a_{Lj}=\frac{e^{\beta \la y,\mathcal Tx_j\ra}}{Z_L(t)},
\]
with
\[
Z_L(t)=e^{\beta \la y,\mathcal Ty\ra}+\sum_{j=1}^{L-1}e^{\beta \la y,\mathcal Tx_j\ra}.
\]
To extract the leading dynamics of $y\in\S^{d-1}$, we introduce 
\[
Z(y)=e^{\beta y\cdot \mathcal Ty}+\sum_{j=1}^{L-1} e^{s_j\beta y\cdot \mathcal T\vec e_1},
\]
and define
\begin{equation}\label{ab-def}
a(y)=\frac{e^{\beta y\cdot \mathcal Ty}}{Z(y)},
\qquad 
b(y)=\frac{1}{Z(y)}\sum_{j=1}^{L-1} s_j e^{s_j\beta y\cdot \mathcal T \vec e_1}.
\end{equation}
\begin{remark}
The functions $Z,a,b$ are the functions $Z_L,a_{LL},\sum_{j=1}^{L-1}a_{Lj}$ evaluated at $x_L=y,x_j=s_j\vec e_1$.
\end{remark}

\begin{lemma}\label{N1N2}
Let $N_1$ denote the number of tokens among $\{x_1,x_2,\dots,x_{L-1}\}$ converging to $\vec e_1$, and $N_2$ the number converging to $-\vec e_1$. Then
\begin{equation}\label{Nab}
\begin{aligned}
Z(y)&=e^{\beta  y\cdot \mathcal Ty}+N_1 e^{\beta y\cdot \mathcal T\vec e_1}+N_2 e^{-\beta y\cdot \mathcal T\vec e_1},\\
a(y)&=\frac{e^{\beta y\cdot \mathcal Ty}}{Z(y)}, 
\qquad 
b(y)=\frac{1}{Z(y)}\bigl(N_1 e^{\beta y\cdot \mathcal T\vec e_1}-N_2 e^{-\beta y\cdot \mathcal T\vec e_1}\bigr).
\end{aligned}
\end{equation}
\end{lemma}

\begin{proof}
This follows directly from the definitions of $a$, $b$, and $Z$.
\end{proof}

\subsection{Equations for the Coordinates of the $L^{th}$ Token}
We decompose
\beq\label{y-com}
y=\sum_{i=1}^d y_i \vec e_i,\qquad y_i(t)=y(t)\cdot\vec e_i.
\eeq
\begin{proposition}\label{small-error}
The dynamics \eqref{y-eq} can be rewritten in the form
\begin{equation}\label{new-yeq}
\partial_t y \;=\; a(y)\,\P_y(Vy) 
\;+\; b(y)\lambda_1\,\P_y(\vec e_1) 
\;+\; \vec g(t),
\end{equation}
where $\vec g(t)$ is an error term satisfying 
\[
|\vec g(t)|\lesssim \eps_{L-1}(t).
\]
\end{proposition}

\begin{proof}
Starting from the equation \eqref{y-eq} and applying the inductive hypothesis 
\eqref{induction}, we obtain the representation \eqref{new-yeq}, where the 
error term $\vec g(t)$ is given explicitly by
\[
\vec g(t)
=\,(a_{LL}-a(y))\P_{y}(Vy)
+\sum_{j=1}^{L-1}\Bigl[a_{Lj}-\frac{s_j e^{s_j\beta y\cdot \mathcal T\vec e_1}}{Z(y)}\Bigr]\P_y(Vx_j)
+\sum_{j=1}^{L-1} a_{Lj}\,\P_y\!\big(V(x_j-s_j\vec e_1)\big).
\]

Since $Z(y)$ and $Z_L$ are uniformly bounded below by a positive constant, it 
remains only to show that 
\[
\bigl| e^{\beta y\cdot \mathcal T x_j}
      - e^{s_j \beta\, y\cdot \mathcal T \vec e_1} \bigr|
\;\longrightarrow\; 0,
\qquad t\to\infty.
\]

For any $X,Y\in[-M,M]$ we have the elementary bound
\[
|e^X - e^Y| \;\le\; e^M\,|X-Y|.
\]
Applying this with 
\[
X=\beta\, y\cdot \mathcal T x_j, 
\qquad 
Y=s_j\beta\, y\cdot \mathcal T \vec e_1,
\]
gives
\[
|e^{\beta y\cdot \mathcal T x_j}
  - e^{s_j \beta y\cdot \mathcal T \vec e_1}|
   \;\le\; e^{\beta |\mathcal T\vec e_1|}\,|\mathcal T|\,\beta\,|y|\,|x_j - s_j\vec e_1|.
\]  This 
completes the proof.
\end{proof}

We now decompose $\vec g(t)$ as
\[
\vec g(t)=\sum_{i=1}^d g_i(t)\vec e_i,\quad g_i(t)=\vec g(t)\cdot \vec e_i
\]
\begin{lemma}\label{coor-wise}
Let $y$ be a solution of \eqref{new-yeq}. Then the first coordinate evolves according to
\begin{equation}\label{y-1}
y_1'(t)
\;=\;
a\bigl(\lambda_1 - Vy\cdot y\bigr)y_1
\;+\;
\lambda_1 b(1-y_1^2)
\;+\;
g_1(t),
\end{equation}
while for each $2\le i\le d$, the coordinate $y_i$ satisfies
\begin{equation}\label{yi-eq}
y_i'(t)
\;=\;
a\bigl(\lambda_i - Vy\cdot y\bigr)y_i
\;-\;
(\lambda_1 b\, y_1)\, y_i
\;+\;
g_i(t).
\end{equation}
\end{lemma}

\begin{proof} The identities \eqref{y-1} and \eqref{yi-eq} then follow directly by taking the inner product of \eqref{new-yeq} with $\vec e_i$. Here we use the identities
\[
\P_y(Vy)\cdot \vec e_i=(\lambda_i-Vy\cdot y) y_i \quad (i\ge 1),\quad \P_y(\vec e_1)\cdot \vec e_1=1-y_1^2,\quad \P_y(\vec e_1)\cdot \vec e_i=-y_1y_i\quad (i\ge 2).
\]
\end{proof}

\subsection{Hierarchical Suppression Mechanism of Transverse Coordinates}\label{reduce-2d}
In this section, we reveal a striking dimensionality--reduction phenomenon in the 
evolution of the final token \(y(t)\in\mathbb{R}^d\).  
Although the dynamics formally reside in a high-dimensional space, the long-time behavior 
is effectively governed by only \emph{a handful of intrinsic degrees of freedom}.  
Put differently, despite the ambient dimension, the token behaves as though it were 
navigating a much lower-dimensional manifold.
To formalize this structure, for each \(\ell\in\{2,\dots,d\}\) we define  
\[
R_\ell(t)=\Biggl(\sum_{j=\ell}^d y_j^2(t)\Biggr)^{1/2},
\qquad
y_{(\ge \ell)}(t)=(y_\ell(t),\dots,y_d(t)).
\]
Thus \(R_\ell(t)\) measures the magnitude of the final \(d-\ell+1\) coordinates of the token:
\[
y(t)
=\Bigl(
y_1(t),\dots,y_{\ell-1}(t),
\underbrace{y_\ell(t),\dots,y_d(t)}_{y_{(\ge \ell)}(t)}
\Bigr).
\]
We further introduce the \emph{error-to-transverse ratio}
\beq\label{ratio}
\kappa_\ell(t)=\frac{\varepsilon_{L-1}(t)}{R_\ell(t)},
\eeq
which compares the upstream residual error---generated by the first \(L-1\) tokens---to the 
transverse radius of the current token. One can easily see that 
\[
\kappa_2(t)\le \kappa_3(t)\le \cdots\le \kappa_d(t)
\]
for all $t\ge 0$. We recall Theorem \ref{sup-main}:

\begin{theorem}[Hierarchical Suppression Mechanism]\label{sup-main1}
Let $y(t)\in\mathbb S^{d-1}$ be the $L$-th token, and let $\varepsilon_{L-1}(t)$ denote 
the residual error contributed by the first $L-1$ tokens.  
For $\ell\in\{2,\dots,d\}$, suppose that,
\begin{equation}\label{con-1}
    \kappa_\ell(t)\longrightarrow 0,
    \qquad (t\to\infty).
\end{equation}
Then exactly one of the following two alternatives occurs:

\medskip
\noindent\emph{(A) $y_\ell$ becomes the dominant transverse direction.}  
\[
y_j(t)\longrightarrow 0,
\qquad\text{for all } j\neq 1,\ell.
\]
In this case the suppression process terminates at level $\ell$.

\medskip
\noindent\emph{(B) $y_\ell$ is suppressed.}
\[
y_\ell(t)\longrightarrow 0.
\]
\end{theorem}

\paragraph{Key Ideas.}
To convey the core idea, let us begin with the simplest nontrivial example:
$L=2$ and $d=3$.  In this case the first token is fixed,
$x_1(t)\equiv \vec e_1$, and we focus on the evolution of the second token
$y=x_2(t)$.  From Lemma~\ref{coor-wise}, the coordinates $y_2$ and $y_3$
satisfy
\beq\label{ran-1209}
y_2' = a(\lambda_2-Vy\cdot y)y_2 - (\lambda_1 b\, y_1)y_2,
\qquad
y_3' = a(\lambda_3-Vy\cdot y)y_3 - (\lambda_1 b\, y_1)y_3.
\eeq  Introducing polar coordinates
$y_2 = R\cos\theta$, $y_3 = R\sin\theta$, we compute
\[
\partial_t 
\begin{pmatrix} y_2\\ y_3\end{pmatrix}
= 
R'(t)
\begin{pmatrix} \cos\theta\\ \sin\theta\end{pmatrix}
+
R\,\theta'(t)
\begin{pmatrix} -\sin\theta\\ \cos\theta\end{pmatrix}.
\]
Taking the dot product with the perpendicular vector $(-y_3,y_2)$ eliminates
the $R'(t)$ term and isolates the angular dynamics:
\[
R^2\theta'
=
\partial_t
\begin{pmatrix} y_2\\ y_3\end{pmatrix}
\cdot
\begin{pmatrix} -y_3\\ y_2\end{pmatrix}
=
a(\lambda_3-\lambda_2)y_2y_3,
\]
and hence
\[
\theta'(t)
=
a(\lambda_3-\lambda_2)\cos\theta\sin\theta.
\]
Thus the two-dimensional dynamics collapse to a single scalar ODE for
$\theta(t)$.  Because $\lambda_3<\lambda_2$, the right-hand side has a fixed
sign away from $\theta\in\{0,\tfrac{\pi}{2},\pi,\tfrac{3\pi}{2}\}$, which
forces $\theta(t)$$ $to move monotonically toward the boundary points.  Consequently,
\[
(y_2(t),y_3(t)) 
\longrightarrow 
\{(\pm R,0),\,(0,\pm R)\},
\]
showing that one of the two coordinates must be suppressed in the long-time
limit.

This illustrates the key \emph{suppression mechanism}: even in higher dimensions, the
dynamics collapse to a low-dimensional angular system in which certain
coordinates are monotonically driven to zero, leaving only one dominant
direction.
\begin{remark}
A major analytical difficulty arises when the transverse radius $R_\ell(t)$ becomes 
small.  In this regime the angular variables develop a $1/R_\ell$ singularity, making it impossible to close the argument 
using the raw coordinates alone.  The quantity $\kappa_\ell(t)$ is introduced 
precisely to measure this \emph{competition}: it compares the strength of the inherited 
error $\varepsilon_{L-1}(t)$ to the geometric scale $R_\ell(t)$ on which the angular 
flow evolves.  The condition $\kappa_\ell(t)\to 0$ ensures that the geometric 
suppression mechanism remains effective even when the transverse radius becomes 
small, allowing the reduced dynamics to overpower any residual perturbation.

Our theorem shows if a coordinate $y_\ell$ continues to occupy a nontrivial portion of the transverse 
space—meaning $R_\ell(t)$ does not decay faster than the upstream error $\eps_{L-1}(t)$—then the 
dynamics force this coordinate to eventually suppress all deeper coordinates 
$y_{\ell+1},\dots,y_d$.  Conversely, if $y_\ell$ decays to zero, the mechanism 
automatically shifts to the next level, where the same dichotomy applies.  
Thus no direction can maintain a marginal presence forever: any coordinate that 
remains consistently filled (up to an $o(1)$ perturbation from earlier tokens) must 
ultimately dominate and eliminate all others.  To the best of our knowledge, this 
hierarchical concentration--versus--suppression phenomenon has not been previously 
justified rigorously in the analysis of token dynamics or related high-dimensional flows.
\end{remark}

\paragraph{Intuition behind the mathematical techniques.}
In the simplified example above, the key idea is to select a vector that is
perpendicular to $(y_2,y_3)$ so that, when we take a dot product with the
evolution equation, the term $-(\lambda_1 b\, y_1) y_i$ in the equation \eqref{ran-1209} cancels out.  
In two dimensions this perpendicular direction is simply $(-y_3, y_2)$.  
In three dimensions, the coordinates $(y_2,y_3,y_4)$ are expressed in spherical
variables, and the appropriate perpendicular directions are given by the
standard orthonormal basis associated with those coordinates. More precisely, we write the vector $(y_2,y_3,y_4)$ in spherical coordinates as
\[
(y_2,y_3,y_4)^\top = R_2(t)\,\mathcal{M}, 
\]
where 
\[
\mathcal{M}
=
\begin{pmatrix}
\cos\varphi_1 \\[0.4em]
\sin\varphi_1\,\cos\varphi_2 \\[0.4em]
\sin\varphi_1\,\sin\varphi_2
\end{pmatrix},
\qquad
R_2(t)=\sqrt{y_2^2+y_3^2+y_4^2}.
\]
A natural perpendicular direction is then 
\[
\partial_{\varphi_1}\mathcal{M}
=
\begin{pmatrix}
-\sin\varphi_1 \\[0.3em]
\cos\varphi_1\,\cos\varphi_2 \\[0.3em]
\cos\varphi_1\,\sin\varphi_2
\end{pmatrix},
\]
and indeed $\mathcal{M}\cdot\partial_{\varphi_1}\mathcal{M}=0$. This same principle extends to general dimension: we choose a direction
orthogonal to $(y_2,\dots,y_d)$ so that the undesirable linear term is
eliminated, allowing us to isolate the key dynamics in the remaining
coordinates.

\begin{proof}[Proof of Theorem~\ref{sup-main}] To simplify the notation, and without loss of generality, we may assume $\ell=2$.  
Our goal is to show that if $\kappa_2(t)\to 0$ as $t\to\infty$, then the dynamics reduce to the two-dimensional subsystem spanned by $(y_1,y_k)$.  
Since all nontrivial behavior occurs only when $d\ge 3$, we henceforth also assume $d\ge 3$.
We introduce standard angular coordinates
$\varphi_1,\ldots,\varphi_{d-2}$ so that
\[
\begin{aligned}
y_2 &= R_2(t)\,\cos\varphi_1, \\
y_3 &= R_2(t)\,\sin\varphi_1\cos\varphi_2, \\
y_4 &= R_2(t)\,\sin\varphi_1\sin\varphi_2\cos\varphi_3, \\
&\;\;\vdots\\
y_{d-1} &= R_2(t)\,\sin\varphi_1\cdots\sin\varphi_{d-3}\cos\varphi_{d-2}, \\
y_d &= R_2(t)\,\sin\varphi_1\cdots\sin\varphi_{d-3}\sin\varphi_{d-2}.
\end{aligned}
\]
We write $\vec y_{\ge 2}(t)=\mathcal M(R_2(t),\varphi)$, where 
$\varphi=(\varphi_1,\ldots,\varphi_{d-2})\in [0,\pi]^{\,d-3}\times[0,2\pi)$.  
It is classical that the spherical coordinate map satisfies  
\[
\partial_{\varphi_i}\mathcal M \;\cdot\; \partial_{\varphi_j}\mathcal M = 0 
\quad (i\neq j), 
\qquad 
\mathcal M\cdot \partial_{\varphi_i}\mathcal M = 0.
\]
By the chain rule, the time derivative of $\mathcal M$ is
\begin{equation}\label{ran-112}
\partial_t \mathcal M 
= R_2'(t)\,\frac{\mathcal M}{R_2}
\;+\;
A\,\partial_t\varphi,
\end{equation}
where the matrix $A\in\R^{(d-1)\times(d-2)}$ has columns
\[
A = \big(\partial_{\varphi_1}\mathcal M,\;
        \partial_{\varphi_2}\mathcal M,\;
        \ldots,\;
        \partial_{\varphi_{d-2}}\mathcal M\big).
\]
Recalling \eqref{yi-eq}, we have
\[
\partial_t \mathcal M
= a(y)\,\diag\{\lambda_2,\lambda_3,\ldots,\lambda_d\}\,\mathcal M
- a(y)\,Vy\cdot y\,\mathcal M
- \lambda_1 b(y)\,y_1\,\mathcal M
+ \vec g_{\ge 2}(t),
\]
where $\vec g_{\ge 2}(t)=(g_2(t),\ldots,g_d(t))$.
Since $\mathcal M\cdot \partial_{\varphi_1}\mathcal M = 0$, we obtain
\begin{equation}\label{ran-113}
\partial_t \mathcal M \cdot \partial_{\varphi_1}\mathcal M
=
a(y)\,\big[\diag\{\lambda_2,\lambda_3,\ldots,\lambda_d\}\big]\mathcal M\cdot\partial_{\varphi_1}\mathcal M
+
\vec g_{\ge 2}(t)\cdot \partial_{\varphi_1}\mathcal M.
\end{equation}

Combining \eqref{ran-112} and \eqref{ran-113}, we get
\[
A\,\partial_t\varphi \,\cdot\, \partial_{\varphi_1}\mathcal M
=
a(y)\,\big[\diag\{\lambda_2,\lambda_3,\ldots,\lambda_d\}\big]\mathcal M\cdot\partial_{\varphi_1}\mathcal M
+
\vec g_{\ge 2}(t)\cdot \partial_{\varphi_1}\mathcal M.
\]
Hence,
\[
A\,\partial_t\varphi \,\cdot\, \partial_{\varphi_1}\mathcal M
=
a(y)\,\big[\diag\{\lambda_2,\lambda_3,\ldots,\lambda_d\}\big]\mathcal M\cdot\partial_{\varphi_1}\mathcal M
+ 
\vec g_{\ge 2}(t)\cdot\partial_{\varphi_1}\mathcal M.
\]
Here we used
\[
A\,\partial_t\varphi \,\cdot\, \partial_{\varphi_1}\mathcal M
=\partial_t\varphi\cdot A^\top\partial_{\varphi_1}\mathcal M
= R_2^2\,\partial_t\varphi_1.
\]
To compute the right-hand side, consider
\[
\sum_{i=2}^d \lambda_i\, y_i\, \partial_{\varphi_1} y_i.
\]
A direct calculation gives
\beq \label{ran-3602}
\sum_{i=2}^d \lambda_i\,y_i\,\partial_{\varphi_1}y_i
=
R_2(t)^2\,\sin\varphi_1\,\cos\varphi_1
\Big(-\lambda_2 + \sum_{i=3}^d \lambda_i\,\alpha_i^2\Big),
\eeq
where the coefficients $\alpha_i$ are given by the standard angular factors
where
\[
\begin{cases}
\alpha_3 = \cos\varphi_2,\\
\alpha_4 = \sin\varphi_2\,\cos\varphi_3,\\
\alpha_5 = \sin\varphi_2\,\sin\varphi_3\,\cos\varphi_4,\\
\vdots\\
\alpha_{d-1} = \Big(\prod_{k=2}^{d-3}\sin\varphi_k\Big)\cos\varphi_{d-2},\\
\alpha_d = \prod_{k=2}^{d-2}\sin\varphi_k.
\end{cases}
\]
%%%%%%%%%%%%%%%%%%%

Since the coefficients $\alpha_3^2,\ldots,\alpha_d^2$ are nonnegative and satisfy
$
\sum_{i=3}^d \alpha_i^2 = 1,
$
the bracket in \eqref{ran-3602} is estimated as
\[
-\lambda_2+\sum_{i=3}^d \lambda_i\alpha_i^2 \le -(\lambda_2-\lambda_3)<0
\]
whenever $\lambda_1>\lambda_2>\cdots>\lambda_d$.
This implies
\[
\begin{aligned}
\sum_{i=2}^d \lambda_i\, y_i\, \pt_{\varphi_1} y_i
    = -c(t) R_2^2 \cos \varphi_1 \sin \varphi_1 .
\end{aligned}
\]
where
\beq\label{c-pos}
c(t)\ge \lambda_2-\lambda_3>0
\eeq
This implies
\begin{equation}\label{ran-1427}
\partial_t \varphi_1
= -a(y)\,c(t)\,\sin\varphi_1\,\cos\varphi_1
+ \frac{\vec g_{\ge 2}(t)\,\cdot\,\partial_{\varphi_1}\mathcal M}
           {R_2(t)^2},
\end{equation}
where $c$ is given in \eqref{c-pos}.
By Proposition \ref{small-error} $|\vec g_{\ge 2}(t)|\lesssim \varepsilon_{L-1}(t)$ and the fact that 
$|\partial_{\varphi_1}\mathcal M|\lesssim R_2(t)$, we obtain
\[
   \frac{\vec g_{\ge 2}(t)\,\cdot\,\partial_{\varphi_1}\mathcal M}
           {R_2(t)^2}\;\lesssim\; \kappa_2(t).
\]
Since $\kappa_2(t)\to 0$ as $t\to\infty$, the forcing term in the equation \eqref{ran-1427} converges to $0$ as $t\to\infty$.
We now show that one of the following two alternatives holds:
\begin{itemize}
\item $(\cos\varphi_1(t),\sin\varphi_1(t)) \to (0,\pm 1)$ as $t\to\infty$,
\item $(\cos\varphi_1(t),\sin\varphi_1(t)) \to (\pm 1,0)$ as $t\to\infty$.
\end{itemize}
\textbf{Claim 1.}
If there exists a sequence $t_n\to\infty$ such that
\[
(\cos\varphi_1(t_n),\sin\varphi_1(t_n))\to(\pm 1,0),
\]
then in fact
\[
(\cos\varphi_1(t),\sin\varphi_1(t))\to(\pm 1,0)
\quad\text{as } t\to\infty.
\]
\textit{Proof of Claim 1.}
Since $\varphi_1(t_n)\in[0,\pi]$, we may assume $\varphi_1(t_n)\to\bar\theta\in[0,\pi]$.
Lemma~\ref{xA} gives $\sin\bar\theta\,\cos\bar\theta=0$, hence
$\bar\theta\in\{0,\frac\pi2,\pi\}$.  
Because $\sin\varphi_1(t_n)\to 0$, we further obtain $\bar\theta\in\{0,\pi\}$.

Choose $\delta_0>0$ sufficiently small so that
\[
\cos(\bar\theta+\delta)\,\sin(\bar\theta+\delta)\ge c_{\delta_0}>0,
\qquad 0<\delta<\delta_0.
\]
We show that $\varphi_1(t)\to\bar\theta$.
Fix $\delta\in(0,\delta_0)$ and define
\[
z(t)=\bar\theta+\delta-\varphi_1(t).
\]
Since $\varphi_1(t_n)\to\bar\theta$, we have $z(t_n)>0$ for all large $n$.
As $\frac{\vec g_{\ge 2}(t)\cdot\partial_{\varphi_1}\mathcal M}{R_2^2}\to 0$, we further assume
\[
\Big|
\frac{\vec g_{\ge 2}(t)\cdot\partial_{\varphi_1}\mathcal M}{R_2^2}
\Big|
\le 0.1 \inf_{y\in\S^d} a(y)(\lambda_2-\lambda_3)c_{\delta_0},
\quad t\ge t_{n_\delta}.
\]
Then for every
\[
T\ge t_{n_\delta}
\quad\text{with}\quad
z(T)=0\quad(\Leftrightarrow\ \varphi_1(T)=\bar\theta+\delta),
\]
we have $z'(T)=-\varphi_1'(T)>0$.  
Lemma~\ref{son-lem} implies $\varphi_1(t)<\bar\theta+\delta$ for all $t\ge t_{n_\delta}$.
A symmetric argument yields $\varphi_1(t)>\bar\theta-\delta$ for all large $t$.
Thus $\varphi_1(t)\to\bar\theta$.  This proves the claim.

\medskip

\textbf{Claim 2.}
Exactly one of the following holds:
\begin{itemize}
\item $(\cos\varphi_1(t),\sin\varphi_1(t))\to(0,\pm 1)$,
\item $(\cos\varphi_1(t),\sin\varphi_1(t))\to(\pm 1,0)$.
\end{itemize}

\textit{Proof of Claim 2.}
Let
\[
\zeta_0=\liminf_{t\to\infty} |\cos\varphi_1(t)|,
\qquad
\zeta_1=\limsup_{t\to\infty} |\cos\varphi_1(t)|.
\]

If $\zeta_0=\zeta_1$, then $\lim_{t\to\infty}|\cos\varphi_1(t)|$ exists.
Using Lemma~\ref{deri-zero} applied to \eqref{ran-1427} and the intermediate value theorem,
the limit must lie in $\{-1,0,1\}$.

Now assume $0\le\zeta_0<\zeta_1\le 1$.
Choose $t_n\to\infty$ such that
\[
|\cos\varphi_1(t_n)|\to\zeta_0,
\qquad
\varphi_1(t_n)\to\bar\theta\in[0,\pi].
\]
Lemma~\ref{deri-zero} again implies $\bar\theta\in\{0,\frac\pi2,\pi\}$, and
$\zeta_0=|\cos\bar\theta|$.

If $\bar\theta\in\{0,\pi\}$, Claim~1 yields $\varphi_1(t)\to\bar\theta$, so
$\zeta_0=\lim|\cos\varphi_1(t)|=1$, contradicting $\zeta_0<\zeta_1$.
Thus $\bar\theta=\frac\pi2$, giving $\zeta_0=0$.

Next choose $t_n\to\infty$ such that $|\cos\varphi_1(t_n)|\to\zeta_1>0$.
Lemma~\ref{deri-zero} gives $\varphi_1(t_n)\to\bar\theta\in\{0,\frac\pi2,\pi\}$ with 
$\zeta_1=|\cos\bar\theta|$.

If $\bar\theta\in\{0,\pi\}$, Claim~1 implies $\varphi_1(t)\to\bar\theta$, hence
$\zeta_1=\lim|\cos\varphi_1(t)|=1=\zeta_0$, a contradiction.
If $\bar\theta=\frac\pi2$, then $\zeta_1=0$, another contradiction.
This proves the claim.
From the arguments above, we have 
\[
y_2(t)=R_2(t)\,\cos\varphi_1(t),
\qquad 
\cos\varphi_1(t)\longrightarrow \pm 1 \ \text{or}\ 0
\quad (t\to\infty).
\]
If $\cos\varphi_1(t)\to 0$, then $y_2(t)\to 0$, and we apply the same polar–coordinate 
analysis to the remaining block $(y_3,\dots,y_d)$, now using $R_3$ and the next angular 
variable.

If instead $\cos\varphi_1(t)\to \pm 1$, then $y_2(t)$ converges to a nonzero limit while 
all other transverse coordinates satisfy $y_j(t)\to 0$ for $j\ge3$.  
In this case $y_2$ is the unique surviving transverse direction, and we set $k=2$.  
The proof is complete.
\end{proof}

\begin{corollary}[Condition for exactly two surviving coordinates]\label{2-survive}
Under the hierarchical suppression mechanism, the long-time dynamics satisfy
\[
y_j(t)\to 0 \qquad (j\neq 1,\ell)
\]
for some unique $\ell\ge 2$  
if and only if there exists an index $\ell\in\{2,\dots,d\}$ such that
\[
\kappa_\ell(t)\to 0
\quad\text{and}\quad 
y_\ell(t)\not\to 0.
\]
In this case the algorithm terminates at level $\ell$, and the only nonvanishing 
coordinates in the limit are $y_1$ and $y_\ell$.
\end{corollary}
\begin{definition}[Dominant--Direction Selection Condition]\label{DDSC}
For a given level $\ell\in\{2,\dots,d\}$, we say that the
\emph{Dominant--Direction Selection Condition} holds if there exists $\ell\in [2,d]$ such that 
\beq\label{DDSC}
\kappa_\ell(t)\to 0
\qquad\text{and}\qquad
y_\ell(t)\not\to 0.
\eeq
Equivalently, this means that
\[
\varepsilon_{L-1}(t)\ \ll\ R_\ell(t),
\]
while the $\ell$-th coordinate itself retains a nonvanishing amplitude.  
When this condition holds, level $\ell$ is the first surviving transverse
direction, and all deeper coordinates $y_{\ell+1},\dots,y_d$ are suppressed by the
hierarchical mechanism.
\end{definition}

\begin{proposition}\label{away-y1}
If there exist $\delta>0$ and $T_\delta>0$ such that 
\[
1-y_1(t)^2\ge \delta \qquad\text{for all } t\ge T_\delta,
\]
then the condition \eqref{DDSC} holds.
\end{proposition}

\begin{proof}
Since $y(t)\in\S^{d-1}$, we have $R_2(t)^2 + y_1(t)^2 = 1$. Hence,
\[
\kappa_2(t)=\frac{\varepsilon_{L-1}(t)}{\sqrt{\,1-y_1(t)^2\,}}.
\]
For all $t\ge T_\delta$, the assumption $1-y_1(t)^2\ge\delta$ implies
\[
\kappa_2(t)\le \delta^{-1/2}\,\varepsilon_{L-1}(t)\to 0
\qquad\text{as } t\to\infty.
\]
If $y_2(t)\not\to 0$, then we choose $\ell=2$. In that case, Theorem~\ref{sup-main} implies that 
$y_j(t)\to 0$ for every $j\ge 3$.
If instead $y_2(t)\to 0$, we proceed to $\kappa_3$. Since $y_2(t)\to 0$, by enlarging $T_\delta$ if necessary we have
\[
R_3(t)^2 = 1-y_1(t)^2 - y_2(t)^2 
\ge \tfrac14 \delta \qquad\text{for all } t\ge T_\delta.
\]
Therefore,
\[
\kappa_3(t)\le 2\,\delta^{-1/2}\,\varepsilon_{L-1}(t)\to 0.
\]
Continuing this argument inductively, we obtain the desired conclusion.
\end{proof}

\begin{theorem} \label{all-zero}
Recall the definition of $N_1, N_2$ in Lemma~\ref{N1N2}. Under the condition \eqref{DDSC}, 
the equation \eqref{y-1} can be written as
\[
\partial_t y_1 = (1-y_1^2) F(y) + U(t),
\]
where $U(t)\to 0$ as $t\to\infty$, and
\beq\label{F-def-1}
\bega
F(t,y)&=\frac{G(t,y)}{Z(t,y)},\\[0.3em]
G(t,y)&=e^{\beta\, y\cdot \mathcal T(t) y}\,(\lambda_1(t)-\lambda_\ell(t))\,y_1
      +\lambda_1(t)\Big[N_1 e^{\beta\, y\cdot\mathcal T(t)\vec e_1}
      -N_2 e^{-\beta\, y\cdot\mathcal T(t)\vec e_1}\Big],\\[0.3em]
Z(t,y)&=e^{\beta\, y\cdot\mathcal T(t) y}
      +N_1 e^{\beta\, y\cdot\mathcal T(t)\vec e_1}
      +N_2 e^{-\beta\, y\cdot\mathcal T(t)\vec e_1}.
\enda
\eeq

When the matrix $\mathcal T = Q^\top K$ has the form \eqref{T-condition}, we obtain
\beq\label{F-def}
\bega
F(t,y_1)&=\frac{G(t,y_1)}{Z(t,y_1)},\\[0.3em]
G(t,y_1)&=e^{\beta\,(m_1(t) + (m_0(t) - m_1(t))y_1^2)}
          \,(\lambda_1(t)-\lambda_\ell(t))\,y_1
       +\lambda_1(t)\Big[N_1 e^{\beta m_0(t) y_1} - N_2 e^{-\beta m_0(t) y_1}\Big],\\[0.3em]
Z(t,y_1)&=e^{\beta\,(m_1(t) + (m_0(t) - m_1(t))y_1^2)}
       +N_1 e^{\beta m_0(t) y_1}
       +N_2 e^{-\beta m_0(t) y_1}.
\enda
\eeq

\end{theorem}

\begin{proof}
From Corollary~\ref{2-survive}, there exists $\ell\in[2,d]$ such that 
$y_j(t)\to 0$ for all $j\neq 1,\ell$. We recall \eqref{y-1}.  
Observe first that
\[
\begin{aligned}
\lambda_1 - Vy\cdot y
&= \lambda_1\sum_{j=1}^d y_j^2 - \sum_{j=1}^d \lambda_j y_j^2
 = \sum_{j=2}^d (\lambda_1 - \lambda_j) y_j^2 \\
&= (\lambda_1 - \lambda_\ell) y_\ell^2
    + \sum_{j\neq \ell} (\lambda_1 - \lambda_j) y_j^2 \\
&= (\lambda_1 - \lambda_\ell) 
    \Bigl[ 1 - y_1^2 - \sum_{j\neq \ell} y_j^2 \Bigr]
    + \sum_{j\neq \ell} (\lambda_1 - \lambda_j) y_j^2 \\
&= (\lambda_1 - \lambda_\ell)(1-y_1^2)
   + \sum_{j\neq \ell}(\lambda_\ell-\lambda_j) y_j^2.
\end{aligned}
\]
Hence the equation for $y_1$ becomes
\[
\partial_t y_1
= (1-y_1^2)\Bigl[a(\lambda_1-\lambda_\ell)y_1 + \lambda_1 b\Bigr]
  + g_1(t)
  + a y_1 \sum_{\substack{j\neq \ell}} (\lambda_\ell - \lambda_j) y_j^2.
\]
By Lemma~\ref{N1N2}, we have
\[
a(\lambda_1-\lambda_\ell) y_1 + \lambda_1 b = F(y).
\]
Define
\[
U(t)
= g_1(t)
  + a y_1 \sum_{\substack{j\neq 1,\ell}} (\lambda_\ell - \lambda_j) y_j^2,
\]
and the desired representation follows.
Finally, note that
\[
y\cdot\mathcal T y = m_1 + (m_0 - m_1)y_1^2,
\qquad
y\cdot\mathcal T\vec e_1 = m_0 y_1,
\]
which yields the stated formulas for $G$ and $Z$.

\end{proof}
\subsection{Proof of Theorem \ref{generalQK}}\label{proofQK}
As in the convention of this paper, we write $y = x_L$.  
By Corollary~\ref{2-survive}, there exists some $\ell \in \{2,\ldots,d\}$ such that 
\[
y_j(t) \to 0 \qquad \text{for all } j \neq 1,\ell.
\]
Since $y(t)$ remains in the compact set $\mathbb S^{d-1}$, we may extract a subsequence $t_n \to \infty$ such that $y(t_n)\to y_\star$ for some limit point $y_\star$.  
We now show that every such limit point must be an equilibrium.

From the above property, the only possibly nonzero components of $y_\star$ are the first and the $\ell$-th.  
Thus $y_\star$ must be of the form
\[
(\alpha_1,0,\ldots,0,\al_\ell,0,\ldots,0),
\qquad \alpha_1^2+\al_\ell^2=1.
\]

Moreover, since
\[
y_\ell^2(t)=1-y_1(t)^2-\sum_{j\neq 1,\ell} y_j(t)^2
  \ge \delta - \sum_{j\neq 1,\ell} y_j(t)^2,
\]
and $y_j(t)\to 0$, we may increase $T_\delta$ so that
\[
y_\ell^2(t)\ge \tfrac 1 4\,\delta \qquad \forall t\ge T_\delta.
\]
Hence $y_\ell(t)$ eventually lies entirely in one of the intervals  
$(-\infty,-\tfrac12\sqrt{\delta})$ or $(\tfrac 12 \sqrt{\delta},\infty)$.  
Without loss of generality, assume $y_\ell(t)>0$ for large $t$.  
Then we may write
\[
y_\ell(t)
  = \sqrt{1-y_1(t)^2 - \sum_{j\neq 1,\ell} y_j(t)^2}
  = \sqrt{1-y_1(t)^2} + y_{\ell,R}(t),
\]
where
\[
y_{\ell,R}(t)
  = \big(1-y_1(t)^2 - \sum_{j\neq 1,\ell} y_j(t)^2\big)^{1/2}
    - \sqrt{\,1 - y_1(t)^2\,}.
\]
Clearly $y_{\ell,R}(t)\to 0$.

Using Theorem~\ref{all-zero} and the fact that $y_j(t)\to 0$ for all $j\neq 1,\ell$, the equation for $y_1$ may be written as
\[
\partial_t y_1
  = P\big(y_1,\sqrt{1-y_1^2}\big) + \widetilde U(t),
\]
where $\widetilde U(t)\to 0$ as $t\to\infty$, and $P$ is a complex-analytic function in both variables.  
Using Lemma~\ref{deri-zero} and the fact that $y_1(t_n)\to\alpha_1$, we obtain
\[
P(\alpha_1,\al_\ell)=0,
\]
so $y_\star$ is an equilibrium point.

Since $\alpha_1^2+\al_\ell^2=1$, we may write
\[
\alpha_1=\cos\theta,\qquad \al_\ell=\sin\theta.
\]
Then $P(\cos\theta,\sin\theta)=0$.  
Because $P$ is complex analytic in both variables, the zeros of this equation are isolated.  
Hence the set of limit points $y_\star$ is isolated.  
Finally, Lemma~\ref{son-lem-3} implies that $y(t)\to y_\star$ as $t\to\infty$.  
This completes the proof.

\section{Instability within the Flow Mechanisms}\label{instability}
The evolution of tokens under a time-dependent $V(t)$ can be intuitively visualized as the motion of a small ball on a hill whose summit itself moves with time, producing much richer mechanism compared to the static case.
To illustrate the idea, let's look at the toy model equation $x(t)\in (-1,1)$ solving the simple ODE
\[
x'(t)=(1-x^2)(x-r(t))
\]
We see that if $r(t)=r_0\in (-1,1)$ is a constant, then no matter where $x(0)$ in $(-1,1)$, we will have $x(t)\to \pm 1$, except at the equilibrium $x=r_0$. We see that $r(t)$ acts like an unstable equilibrium. The physical analogy is as follows:
\begin{itemize}
    \item \textbf{Rapidly moving hilltop.} 
    If the hilltop $x=r(t)$ moves quickly from left $-1$ to right $1$, the ball (initial position $x(0)$) cannot roll off either side; 
  so $x(t)$ remains trapped near an intermediate orientation, forming an \emph{interior cluster}.
    
    \item \textbf{Fixed hill.}
    If the hill $r(t)=r_0$ is fixed, the ball eventually rolls down one side $\pm 1$.
\end{itemize}
This ``moving-hill'' analogy thus captures the physical mechanism behind the emergence of metastable or oscillatory interior clusters on the sphere.
\paragraph{A toy model.}
To be more precise, we now assume simple scenarios first. 
First we assume $Q^\top K=I_d$ for simplicity. To make things even more simple, we assume $x_1(t)\equiv \vec e_1$. We have 
\[
x_2(t)=x(t)\vec e_1+y(t)\vec e_2,
\]
so that $(x(t),y(t))$ solves
\[
\pt_t x_2=a\,\P_{x_2}(Vx_2)+b\lambda_1 \P_{x_2}(\vec e_1),
\]
As in the proof of Theorem \ref{toy-case}, the $x$–equation reduces to
\beq\label{toy-1}
\begin{aligned}
x'(t)&=\frac{ (1-x^2)}{e^{\beta}+e^{\beta x}}
\,\lw\{e^\beta(\lambda_1(t)-\lambda_2(t))x+\lambda_1(t) e^{\beta x}\rw\}.
\end{aligned}
\eeq
We specify 
\[
\lambda_1(t)=1
\]
for all $t\ge 0$.
Let 
\[
g(t,x)=e^\beta(1-\lambda_2(t))x+ e^{\beta x},\quad h(x)=\frac{ 1}{e^{\beta}+e^{\beta x}}.
\]
We are left to study 
\[
x'(t)=(1-x^2) h(x)g(t,x)
\]
with $\lambda_2(t)\in (0,1)$. We note that $g(t,1)=e^\beta(1-\lambda_2)+e^\beta>0$ and $g(t,-1)=e^{-\beta}-e^\beta(1-\lambda_2)$. Hence, as long as 
\[
1-\lambda_2(t)>e^{-2\beta}
\]
for all $t\ge 0$, we will have $g(t,-1)<0$. This garantee that there exists $r(t)\in (-1,1)$ such that
\[
g(t,r(t))=0\quad\forall t\ge 0
\]
At the same time, since \[
\pt_x g(t,x)=e^\beta(1-\lambda_2)+\beta e^{\beta x}\ge e^{-\beta}>0,
\] the map $t\to r(t)$ is well-defined. We can also see that from $g(t,r)=0$, we have $r(t)\in (-1,0)$.
Moreover, we note that from $r(t)$, we can recover $\lambda_2(t)$ as follows
\[
|r(t)|e^{\beta |r(t)|}=\frac{e^{-\beta}}{1-\lambda_2(t)}
\]
If $\lambda_2(t)\to\lambda_{2,\infty}\in (0,1)$, we see that 
\[
r(t)\to r_\infty \in(-1,0)
\]
where $r_\infty$ solves 
\[
|r_\infty|\,e^{\beta |r_\infty|}
= \frac{e^{-\beta}}{1-\lambda_{2,\infty}}.
\]
This observation motivates our main Theorem~\ref{almost}, which is formulated precisely to handle models of the form \eqref{toy-1} that arise naturally from token dynamics.
\begin{theorem}\label{almost}
Let $x:[0,\infty)\to(-1,1)$ be a solution of the differential equation
\[
x'(t)=(1-x(t)^2)\,h(x(t))\,g(t,x(t)) + w(t,x(t)).
\]
Assume the following:
\begin{itemize}
\item The perturbation $w(t,x)$ satisfies
\begin{equation}\label{0limit}
\sup_{x\in[a,b]} |w(t,x)| \longrightarrow 0
\quad\text{as } t\to\infty,
\end{equation}
for any $a,b\in(-1,1)$, and
\[
|w(t,x_1)-w(t,x_2)| \le W(t)\,|x_1-x_2|,
\]
where $W(t)\to 0$ as $t\to\infty$.

\item There exists a constant $c_0>0$ such that
\[
h(x)\ge c_0,
\qquad
\inf_{t\ge0}\partial_x g(t,x)\ge c_0,
\]
for all $x\in\mathbb R$.

\item The function $z\mapsto h(z)^{-1}$ is entire.

\item For each $t\ge0$, there exists a root $r(t)\in(-1,1)$ of
\[
g(t,r(t))=0,
\]
satisfying
\[
r(t)\in[\alpha,\beta]\quad\text{for all }t\ge0,
\qquad
\lim_{t\to\infty} r'(t)=0,
\]
for some $\alpha,\beta\in(-1,1)$.
\end{itemize}
Then there exists \emph{at most one} trajectory $x_\star(t)\in(-1,1)$ that does not have $\pm1$ as a limit point as $t\to\infty$. Moreover, if the condition \eqref{0limit} is replaced by
\beq\label{0limit-1}
\sup_{x\in [-1,1]}|w(t,x)|\to 0\quad\text{as}\quad t\to\infty
\eeq
then
for any initial condition $x(0)\neq x_\star(0)$, the corresponding solution satisfies
\[
\lim_{t\to\infty} x(t)\in\{-1,1\}.
\]
\end{theorem}

\begin{remark}
    Theorem~\ref{almost} provides a general dynamical principle for one-dimensional
token flows of the form \eqref{toy-1}.  
Such equations arise naturally after projecting the high-dimensional 
token dynamics \emph{onto the dominant direction}, where the remaining degrees of
freedom contribute through a slowly varying ``effective root'' $r(t)$.
Informally, Theorem~\ref{almost} states that once all the remaining
(non-dominant) directions of the token dynamics are controlled inside a
bounded interval with only mild fluctuations, the evolution along the
\emph{dominant direction} becomes essentially one-dimensional and highly
rigid.  Under these mild conditions, the reduced flow collapses almost
completely into one of the two extremal states $\{-1,1\}$: there is at
most \emph{one} exceptional trajectory that does not converge to an
endpoint, while every other initialization converges decisively to either
$+1$ or $-1$.

\end{remark}

\begin{proof} If $x(t)$ does not admit $\pm 1$ as a limit point as $t\to\infty$, then there exists $a,b\in (-1,1)$ and $T>0$ such that $x(t)\in [a,b]$ for all $t\ge T$.~\\
We will show that there cannot be two distinct flows $x_1,x_2$ satisfying the above property. 
We define $u(t)=\Phi(x(t))$, where 
\[
\Phi(x)=\int_0^x \frac{1}{(1-w^2)h(w)}\,dw.
\]
The function $\Phi:(-1,1)\to\R$ has the following properties:
\begin{itemize}
\item $\Phi'(x)=\frac{1}{(1-x^2)h(x)}$ for all $x\in(-1,1)$.
\item $\Phi$ is an increasing bijection from $(-1,1)$ to $\R$, with $\Phi(x)\to -\infty$ as $x\to(-1)^+$ and $\Phi(x)\to\infty$ as $x\to 1^{-}$, and $\Phi'(x)>0$.
\end{itemize}

Now for $u=\Phi(x)$, we can write 
\[
x=f(u), \quad f=\Phi^{-1}, \quad f:\R\to (-1,1).
\]
The equation for $u$ reads
\[
u'(t)=g(t,f(u)) + \frac{w(t,f(u))}{(1-x^2)h(x)}.
\]
Now assume that there are two distinct flows $x_1(t),x_2(t)$ such that
\[
x_1(t),x_2(t)\in [a,b] \quad \forall t\ge T.
\]
Define $u_1=\Phi(x_1)$, $u_2=\Phi(x_2)$, and
\[
H(x)=\frac{1}{(1-x^2)h(x)}.
\]
We see that there exists $C_{a,b}>0$ such that
\[
0<H(x)\le C_{a,b} \quad \forall x\in[a,b].
\]
We have 
\[
\begin{aligned}
\frac{d}{dt}[u_1(t)-u_2(t)]
&=g(t,f(u_1))-g(t,f(u_2))+w(t,x_1)\{H(x_1)-H(x_2)\}+H(x_2)(w(t,x_1)-w(t,x_2))\\
&=\partial_x g(t,f(\xi_u))[f(u_1)-f(u_2)]
  +w(t,x_1)\{H(x_1)-H(x_2)\}+H(x_2)(w(t,x_1)-w(t,x_2))\\
&=\partial_x g(t,f(\xi_u)) f'(\bar\xi_u)(u_1-u_2)
   +w(t,x_1)\{H(x_1)-H(x_2)\}+H(x_2)(w(t,x_1)-w(t,x_2)),\\
\end{aligned}
\]
for some $\xi_u,\bar\xi_u\in (u_1(t),u_2(t))$. Now since $f=\Phi^{-1}$, we have 
\[
f'(u)=\frac{1}{\Phi'(f(u))}=[1-f(u)^2]h(f(u)).
\]
Hence we get
\[
\begin{aligned}
\frac{d}{dt}[u_1(t)-u_2(t)]
&=\partial_x g(t,f(\xi_u))[1-f(\bar\xi_u)^2]h(f(\bar\xi_u))[u_1(t)-u_2(t)]\\
&\quad + w(t,x_1)\{H(x_1)-H(x_2)\}+H(x_2)(w(t,x_1)-w(t,x_2)).
\end{aligned}
\]
Letting $\xi_x(t)=f(\xi_u)$ and $\bar\xi_x=f(\bar\xi_u)$, we have $\xi_x,\bar\xi_x\in (x_1(t),x_2(t))$, and 
\[\bega
\frac{d}{dt}[u_1(t)-u_2(t)]
&=\partial_x g(t,\xi_x)[1-\bar\xi_x^2]h(\bar\xi_x)[u_1(t)-u_2(t)]\\
&+ w(t,x_1)\{H(x_1)-H(x_2)\}+H(x_2)(w(t,x_1)-w(t,x_2)).
\enda
\]
Since $x_1(t),x_2(t)\in[a,b]$ for $t\ge T$, there exists $c_0>0$ such that 
\[
\begin{cases}
1-\bar\xi_x(t)^2\ge c_0,\\
|H(x_1)-H(x_2)|\le c_0|u_1-u_2|\\
\end{cases}
\]
We define 
\[
z(t)=(u_1(t)-u_2(t))^2.
\]
Then, for $T$ large and $t\ge T$, we get
\[
z'(t)\ge \lw(c_0-|w(t)|-\sup_{x_2\in [a,b]}|H(x_2)| W(t)\rw) z(t)\ge 0.1 c_0 z(t),
\]
for some constant $c_0$. 
Since $u_1,u_2$ are bounded for $t\ge T$, this is a contradiction. 
This implies that there is at most one solution $x_\star$ that does not have $\pm 1$ as a limit point. 
This completes the proof.
Now we show that under assumption~\eqref{0limit-1}, if $\pm1$ is a limit point of
$\{x(t)\}_{t>0}$, then $x(t)\to\pm1$ as $t\to\infty$.
It suffices to consider the case where $+1$ is a limit point, as the case
$-1$ is treated analogously.

Assume therefore that there exists a sequence $t_n\to\infty$ such that
$x(t_n)\to1$.
We claim that $x(t)\to1$ as $t\to\infty$.
By compactness, passing to a subsequence if necessary, we may assume
\[
r(t_n)\to r_\infty\in[\alpha,\beta]\subset(-1,1).
\]

Fix $\varepsilon\in(0,1-\beta)$ and define
\[
y(t):=x(t)-r(t)-\varepsilon.
\]
Since $x(t_n)\to1$ and $r(t_n)\to r_\infty$, we have
\[
\lim_{n\to\infty} y(t_n)=1-r_\infty-\varepsilon
\ge 1-\beta-\varepsilon>0.
\]
Hence there exists $n_\varepsilon$ such that
\[
y(t_n)>0\quad\text{for all }n\ge n_\varepsilon.
\]

Using  \eqref{0limit-1} and $r'(t)\to0$ as $t\to\infty$, we may increase $n_\varepsilon$
so that
\begin{equation}\label{ran-huh}
\sup_{x\in [-1,1]}|w(t,x)|+|r'(t)|
\le
0.8\,c_0^2\,\varepsilon
\inf_{r\in[\alpha,\beta]}\bigl(1-(r+\varepsilon)^2\bigr),
\quad
\forall\,t\ge t_{n_\varepsilon}.
\end{equation}

We now show that
\[
y'(T)>0
\quad\text{for all }T\ge t_{n_\varepsilon}\text{ such that }y(T)=0.
\]
Indeed, if $y(T)=0$, then $x(T)=r(T)+\varepsilon$, and thus
\[
\begin{aligned}
y'(T)
&=(1-x^2)h(x)g(T,x)+w(T,x(T))-r'(T)\\
&=[1-(r(T)+\varepsilon)^2]\,
h(x(T))\,g(T,r(T)+\varepsilon)
+w(T,x(T))-r'(T)\\
&\ge
c_0\,[1-(r(T)+\varepsilon)^2]\,
g(T,r(T)+\varepsilon)
-|w(T,x(T))|-|r'(T)|.
\end{aligned}
\]
Using $g(T,r(T))=0$ and the mean value theorem, we obtain
\[
g(T,r(T)+\varepsilon)
=
\partial_x g(T,\xi)\,\varepsilon
\ge c_0\varepsilon,
\]
for some $\xi\in(r(T),r(T)+\varepsilon)$.
Therefore,
\[
\begin{aligned}
y'(T)
&\ge
c_0^2\,\varepsilon\,[1-(r(T)+\varepsilon)^2]
-\sup_{x\in [-1,1]}|w(T,x)|-|r'(T)|\\
&\ge
c_0^2\,\varepsilon
\inf_{r\in[\alpha,\beta]}\bigl(1-(r+\varepsilon)^2\bigr)
-\sup_{x\in [-1,1]}|w(T)|-|r'(T)|.
\end{aligned}
\]
By~\eqref{ran-huh}, we conclude that $y'(T)>0$.

By Lemma~\ref{son-lem}, it follows that
\[
y(t)>0\quad\text{for all }t\ge t_{n_\varepsilon},
\]
which implies
\begin{equation}\label{ran-1217}
x(t)-r(t)>\varepsilon
\quad\text{for all }t\ge t_{n_\varepsilon}.
\end{equation}

Next, we write
\[
\begin{aligned}
x'(t)
&=(1-x^2)h(x)g(t,x)+w(t)\\
&=(1-x^2)h(x)\bigl[g(t,r(t))
+\partial_x g(t,\xi)(x-r(t))\bigr]+w(t)\\
&=(1-x^2)h(x)\,\partial_x g(t,\xi)\,(x-r(t))+w(t),
\end{aligned}
\]
for some $\xi(t)\in(r(t),x(t))$.
Combining this with~\eqref{ran-1217}, we obtain for all $t\ge t_{n_\varepsilon}$,
\[
x'(t)\ge c_0^2\,\varepsilon\,(1-x^2)-\sup_{x\in [-1,1]}|w(t,x)|.
\]

Consequently,
\[
\begin{aligned}
\frac{d}{dt}(1-x)
&=-x'(t)\\
&\le
-c_0^2\,\varepsilon\,(1+x)(1-x)+\sup_{x\in [-1,1]}|w(t,x)|\\
&\le
-c_0^2\,\varepsilon\,(1+r(t)+\varepsilon)(1-x)+\sup_{x\in [-1,1]}|w(t,x)|\\
&\le
-c_0^2\,\varepsilon\,(1+\alpha+\varepsilon)(1-x)+\sup_{x\in [-1,1]}|w(t,x)|.
\end{aligned}
\]
By Lemma~\ref{Trinh-lem}, this yields $x(t)\to1$ as $t\to\infty$.

\end{proof}
In the next proposition, we show that our theorem is sharp and cannot be strengthened to assert that \emph{all} flows converge to $\pm \vec e_1$. 
We provide an explicit example in the simplified setting $L=d=2$ with the first token equal to $\vec e_1$, and exhibit a trajectory $x_\star$ whose limit is neither $+\vec e_1$ nor $-\vec e_1$.

\begin{proposition} \label{trap-case}
Take $\alpha\in (-1,0)$. Then there exists an ODE of the form \eqref{toy-1} with suitable $\beta$, with $\lambda_1(t)\equiv 1>\lambda_2(t)$ and $\lambda_2(t)\to \lambda_{2,\infty}\in (0,1)$ at an exponential rate, such that $x(0)=\alpha$ and 
\[
\lim_{t\to\infty} x(t)=\alpha.
\]
In particular, the corresponding trajectory $x(t)$ converges to the interior point $\alpha\notin\{-1,1\}$.
\end{proposition}

\begin{proof} 
We take $\lambda_1(t)=1$. Fix a small $\varepsilon_0\in (0,1)$ and set
\[
x(t)=\alpha+\varepsilon_0 e^{-t}\sin t.
\]
Then $x(t)\to \alpha$ as $t\to\infty$, and for all $t\ge 0$,
\[
x(t)\in[\alpha-\varepsilon_0,\alpha+\varepsilon_0]\subset(-1,0).
\]
Next we determine $\lambda_2(t)$. From the equation
\beq\label{ran-2104}
x'(t)=h(x)\bigl[e^\beta(1-\lambda_2(t))x+e^{\beta x}\bigr], 
\qquad 
h(x)=\frac{1-x^2}{e^\beta+e^{\beta x}},
\eeq
we obtain
\[
e^\beta(1-\lambda_2)x+e^{\beta x}
=\frac{e^\beta+e^{\beta x}}{1-x^2}\,x'(t).
\]
Solving for $\lambda_2$ gives
\[
\lambda_2
= 1 
+ \frac{e^{\beta(x-1)}}{x}
- \frac{(1+e^{\beta(x-1)})\,x'}{x(1-x^2)}.
\]
Substituting the explicit form of $x(t)$ yields
\[
\lambda_2(t)
= 1
+ \frac{e^{\beta(\alpha+\varepsilon_0 e^{-t}\sin t - 1)}}
       {\alpha+\varepsilon_0 e^{-t}\sin t}
- \frac{\bigl(1+e^{\beta(\alpha+\varepsilon_0 e^{-t}\sin t - 1)}\bigr)
        \,\varepsilon_0 e^{-t}(\cos t-\sin t)}
       {(\alpha+\varepsilon_0 e^{-t}\sin t)\,
        \bigl(1-(\alpha+\varepsilon_0 e^{-t}\sin t)^2\bigr)}.
\]
As $t\to\infty$, we have $x(t)\to \alpha$ and $x'(t)\to 0$, hence
\beq\label{ran-2105}
\lambda_2(t)\to \lambda_{2,\infty}
=1+\frac{e^{\beta(\alpha-1)}}{\alpha}\in (0,1),
\eeq
provided that $\alpha\in(-1,0)$ and 
\[
\beta>\frac{\ln(-\alpha)}{\alpha-1}.
\]
This completes the proof.
\end{proof}

\begin{remark}
We have taken the first token to be $\vec e_1$, and here $\alpha=x_2(0)\cdot \vec e_1<0$.  
Thus the initial position of the second token lies on the opposite hemisphere from $\vec e_1$ (see Table \ref{tab:prior_work_comparison}).
\end{remark}

\begin{remark}
By Theorem~\ref{almost}, the solution of \eqref{ran-2104} with $\lambda_2(t)$ given by \eqref{ran-2105} converges to $\pm 1$ for every initial condition $x(0)\neq \alpha$.  
The function $\lambda_2(t)$ itself satisfies the asymptotic behavior
\[
\lambda_2(t)
=\lambda_{2,\infty}
+O(e^{-t}\cos t)
+O(e^{-t}\sin t),
\]
so $\lambda_2(t)$ approaches its limit exponentially with a decaying oscillation.
\end{remark}

\section{Time-dependent $Q(t),V(t),K(t)$}\label{full-proof}
In this section, we give the proof of Theorem~\ref{time-dependent-main}.  
If $y(t_n)\to \pm \vec e_1$ for some sequence $t_n\to\infty$, then the theorem follows immediately.  
If this does not occur, then there exist $\delta_0>0$ and $T>0$ such that 
\[
1 - y_1(t)^2 \ge \delta_0 \qquad \forall\, t\ge T.
\]
By Theorem \ref{all-zero} and Proposition \ref{away-y1}, the dynamics for 
$y_1 = y\cdot \vec e_1$ reduce to
\[
\partial_t y_1 = (1 - y_1^2)\,F(t,y_1) + U(t,y_1),
\]
where $U(t)\to 0$ as $t\to\infty$, and $F(t,y_1)$ is defined in~\eqref{F-def}:
\[
\begin{cases}
F(t,y_1) = \dfrac{G(t,y_1)}{Z(t,y_1)},\\[0.8em]
G(t,y_1)
= e^{\beta(m_1(t) + (m_0(t)-m_1(t)) y_1^2)}(\lambda_1 - \lambda_k)\,y_1
  + \lambda_1\!\left[N_1 e^{\beta m_0(t) y_1} - N_2 e^{-\beta m_0(t) y_1}\right],\\[0.6em]
Z(t,y_1)
= e^{\beta(m_1(t) + (m_0(t)-m_1(t)) y_1^2)} 
  + N_1 e^{\beta m_0(t) y_1}
  + N_2 e^{-\beta m_0(t) y_1}\\
U(t,y_1)
= g_1(t)
  + a(t,y_1) y_1 \sum_{\substack{j\neq 1,\ell}} (\lambda_\ell - \lambda_j) y_j^2\\
\end{cases}
\]
By a direct calculation, we have 
\[
\pt_{y_1}G(t,y_1)
= e^{\beta y\cdot\mathcal T y}(\lambda_1-\lambda_k)
  \Big[1 + 2\beta (m_0 - m_1) y_1^2\Big]
  + \beta m_0\Big[\lambda_1 N_1 e^{\beta m_0 y_1}
  + \lambda_1 N_2 e^{-\beta m_0 y_1}\Big],
\]
First, we will show that
\begin{equation}\label{G-prime}
\begin{aligned}
G'(y_1)
&\ge \lambda_1 \beta m_0 L
   + (\lambda_1 - \lambda_k)
     \min\!\Big\{1,\, 1 - 2\beta(m_1 - m_0)\Big\}.
\end{aligned}
\end{equation}
If $m_0 \ge m_1$, then clearly
\[
G'(y_1) \ge \lambda_1 \beta m_0 L + \lambda_1 - \lambda_k > 0.
\]
If $m_0 < m_1$, then
\[
1 + 2\beta(m_0 - m_1) y_1^2
= 1 - 2\beta(m_1 - m_0) y_1^2
\ge 1 - 2\beta(m_1 - m_0),
\]
completing the proof of \eqref{G-prime}.
 Fix any $t\ge 0$, we see that the map $y_1\mapsto G(t,y_1)$ is strictly increasing on $\mathbb{R}$, thanks to \eqref{assumption1} and the bound \eqref{G-prime}. Define the limiting function
\[
G_\infty(y_1) = \lim_{t\to\infty} G(t,y_1).
\]
Our strategy is to apply Theorem~\ref{almost}.  
We split the analysis into two cases:
\begin{itemize}
\item $G_\infty(-1)>0$ or $G_\infty(1)<0$\quad (monotone case),
\item $G_\infty(-1)<0<G_\infty(1)$ \quad (sign-changing case).
\end{itemize}

\medskip
\noindent\textbf{Case 1: $G_\infty(-1)>0$ or $G_\infty(1)<0$.}  
Assume first that $G_\infty(-1)>0$ (the case $G_\infty(1)<0$ is symmetric).  
For all sufficiently large $t$,
\[
G(t,y_1)\ge G(t,-1) \ge 0.9\,G_\infty(-1) > 0.
\]
Thus, for $t\ge T$ with $T$ large enough,
\[
y_1'(t)
\ge (1 - y_1(t)^2)\,0.9\,G_\infty(-1) - |U(t)|
\ge 0.9\,\delta_0\,G_\infty(-1) - |U(t)|.
\]
Since $U(t)\to 0$, we eventually have $y_1'(t)>0$.  
Hence $y_1(t)$ is strictly increasing for large $t$, and as a bounded monotone function on $(-1,1)$ it must converge:
\[
y_1(t) \to \alpha \in [-1,1].
\]
Taking the limit in the equation yields
\[
(1 - \alpha^2)\,G_\infty(\alpha) = 0,
\]
so $\alpha \in \{-1,1\}$.  
Thus $y_1(t)\to\pm 1$, completing Case~1.

\medskip
\noindent\textbf{Case 2: $G_\infty(-1)<0<G_\infty(1)$.}  
In this regime, $G(t,-1)<0<G(t,1)$ for all sufficiently large $t$.  
Fix $T>0$ so that this holds for all $t\ge T$.  
By strict monotonicity of $G(t,\cdot)$, for each $t\ge T$ there exists a unique $r(t)\in(-1,1)$ such that
\[
G(t,r(t)) = 0.
\]

We show that $r(t)\to r_\infty$ for some $r_\infty\in (-1,1)$ with $G_\infty(r_\infty)=0$, and that $r'(t)\to 0$.

\smallskip
\underline{Step 1: Convergence of $r(t)$.}  
Let $t_n\to\infty$ be arbitrary.  
Since $r(t_n)$ is bounded, extract a subsequence (not relabeled) such that $r(t_n)\to\alpha$.  
Passing to the limit in $G(t_n,r(t_n))=0$ yields
\[
G_\infty(\alpha)=0.
\]
By strict monotonicity of $G_\infty$, the root is unique, hence $\alpha=r_\infty$.  
Therefore $r(t)\to r_\infty$.

\smallskip
\underline{Step 2: Decay of $r'(t)$.}  
Differentiating $G(t,r(t))=0$ gives
\[
\partial_t G(t,r(t))
+ \partial_{y_1}G(t,r(t)) \, r'(t) = 0,
\]
so
\[
r'(t)
= -\frac{\partial_t G(t,r(t))}{\partial_{y_1}G(t,r(t))}.
\]
From~\eqref{G-prime}, $\partial_{y_1}G(t,r)\ge\lambda_1\beta m_0>0$.  
Thus
\[
|r'(t)|
\lesssim |\partial_t G(t,r(t))|
\lesssim |\lambda_1'(t)| + |\lambda_k'(t)| \to 0.
\]

\smallskip
\underline{Step 3: Verification of the monotonicity condition in Theorem~\ref{almost}.}  
We must show that
\[
\inf_{t\ge 0}
\frac{\mathrm{d}}{\mathrm{d}y_1}
\left[
\frac{Z_\infty(y_1)}{Z(t,y_1)}\,G(t,y_1)
\right] > 0.
\]
Compute
\[
\frac{\mathrm{d}}{\mathrm{d}y_1}
\left(\frac{Z_\infty}{Z}G\right)
=
\frac{\mathrm{d}}{\mathrm{d}y_1}\!\left(\frac{Z_\infty}{Z}\right) G
+\frac{Z_\infty}{Z}\,\partial_{y_1}G
\ge 
c_0\!\left(\partial_{y_1}G 
-
G\left|\frac{\mathrm{d}}{\mathrm{d}y_1}\!\left(\frac{Z_\infty}{Z}\right)\right|\right).
\]
Since $\partial_{y_1}G$ is uniformly bounded below, it suffices to show
\[
\frac{\mathrm{d}}{\mathrm{d}y_1}\!\left(\frac{Z_\infty}{Z}\right)
\to 0
\quad\text{uniformly for } y_1\in[-1,1].
\]
Indeed,
\[
\begin{aligned}
\left|\partial_{y_1}\!\left(\frac{Z_\infty}{Z}\right)\right|
&\lesssim |\partial_{y_1}Z_\infty|\,|Z-Z_\infty|
  + |Z_\infty|\,|\partial_{y_1}(Z-Z_\infty)| \\
&\lesssim |Z-Z_\infty|
  + |\partial_{y_1}(Z-Z_\infty)| \\
&\lesssim |m_0(t)-m_{0,\infty}|
       + |m_1(t)-m_{1,\infty}|.
\end{aligned}
\]
Since $m_0(t)\to m_{0,\infty}$ and $m_1(t)\to m_{1,\infty}$, this quantity tends to zero, completing the verification.

The proof is complete.
\section{Technical Lemmas}
In this subsection we collect several technical lemmas that will be used
throughout the paper.  These results serve as analytic tools inspired by
classical methods in mathematical analysis.  Readers interested only in the
high-level flow of the argument may safely skip this section on a first pass
and return to it as needed.
\begin{lemma}\label{son-lem} Let \( g : [t_0, \infty) \to \mathbb{R} \) be a differentiable function with a continuous derivative \( g' \). Suppose \( g(t_0) > \delta \) for some \( \delta \in \mathbb{R} \). If \( g'(T) > 0 \) for every $T$ in the set \[ \{T\ge t_0:\quad g(T)=\delta\} \] then \[ g(t) > \delta \qquad \text{for all } t \ge t_0. \] \end{lemma}

\begin{proof} Suppose, for contradiction, that there exists $t>t_0$ with $g(t)\le \delta$. Since $g(t_0)>\delta$, there exists a time $T>t_0$ which is the smallest time such that $g(T)=\delta$. We then have $g'(T)>0$. By continuity of $g'$, there exists $\varepsilon_0>0$ such that \[ g'(t)>0 \qquad \text{for all } t\in (T-\varepsilon_0,T+\varepsilon_0). \] In particular, for $t\in (T-\varepsilon_0,T)$ we have $g(t)<g(T)=\delta$. By continuity of $g$, this implies there exists $T_1\in (t_0,T)$ with $g(T_1)=\delta$, contradicting the minimality of $T$. Hence no such $T$ can exist, and we conclude \[ g(t)>\delta \qquad \text{for all } t\ge t_0. \] \end{proof}
\begin{lemma}[Limit-point principle for scalar residual dynamics]\label{deri-zero}
Let $c:[0,\infty)\to\mathbb R_+$ be a smooth function satisfying the uniform
bounds
\[
0 < c_\infty \;\le\; c(t) \;\le\; C_\infty < \infty 
\qquad \text{for all } t\ge 0.
\]
Consider the scalar evolution
\begin{equation}\label{xA}
x'(t) = c(t)\,\mathcal A(x(t)) + u(t),
\end{equation}
where $u(t)\to 0$ as $t\to\infty$ and $\mathcal A:\mathbb R\to\mathbb R$ is
continuous.  Then any accumulation point of $x(t)$ must be a zero of the drift:
if $x(t_n)\to \alpha$ along some sequence $t_n\to\infty$, we must have
\[
\mathcal A(\alpha)=0.
\]
\end{lemma}

\begin{proof}
Assume that $\mathcal A(\al)\neq 0$, then we have $\mathcal A(\al)>0$ or $\mathcal A(\al)<0$.~\\
\textbf{Case 1:} $\mathcal A(\al)>0$~\\
Since $\mathcal A$ is continuous, we can find $\delta>0$ such that $\mathcal A(x)>0$ for all $x\in [\al-9\delta,\al+9\delta]$. We will first show that there exists $T_\delta>0$ such that 
\beq\label{ran-902}
x(t)>\al-\delta\quad\text{for all}\quad t\ge T_\delta.
\eeq
To this end, we consider 
\[
z(t)=x(t)-\al+\delta.
\]
Since $x(t_n)\to \al$ as $n\to\infty$, we can find $n_0>0$ such that $z(t_n)>0$ for all $n\ge n_0$. Moreover, since $u(t)\to 0$ as $t\to\infty$, we may further assume 
\beq\label{ran-901}
|u(t)|\le 0.1 c_\infty \mathcal A(\al-\delta)\quad\text{for all}\quad t\ge t_{n_0}.
\eeq
Now as soon as $z(T)=0$ for some $T\ge t_{n_0}$, we have $x(T)=\al-\delta$, and thus
\[\begin{aligned} 
z'(T)&=x'(T)=c(T)\mathcal A(\al-\delta)+u(T)\\
&\ge c_\infty \mathcal A(\al-\delta)-0.1 c_\infty \mathcal A(\al-\delta)>0
\end{aligned} 
\]
thanks to \eqref{ran-901}. By Lemma \ref{son-lem}, we obtain \eqref{ran-902}. Next we show that for all $\delta>0$ sufficiently small and for all $T>0$ large, there exists $t=t(\delta,T)>0$ such that $|x(t)-\al|<\delta$ and $|x'(t)|<\delta$. If this is true, then we can find a sequence $\tau_n\to\infty$ such that $x(\tau_n)\to\al$ and $x'(\tau_n)\to 0$. Passing $\tau_n\to\infty$ in \eqref{xA}, we have $\mathcal A(\al)=0$. To this end, we consider two cases:~\\
\textbf{Subcase 1:} There exists $T>T_\delta$ such that $x(t)\le \al+\delta$ for all $t\ge T$.
In this case, we have $t\in [T,T+2]$ such that $x(T+2)-x(T)=2x'(t)$. Hence
\[
|x'(t)|=\frac{|x(T+2)-x(T)|}{2}\le \frac{(\al+\delta)-(\al-\delta)}{2}=\delta.
\]
\textbf{Subcase 2:} For all $T>T_\delta$, there exists $t>T$ such that $x(t)>\al+\delta$. In this case, we can find $s_1>T_\delta$ such that
\[
x(s_1)>\al+\delta.
\]
Since $x(t_n)\to \al$ as $n\to\infty$, we can find $t_n>s_1$ such that 
\[
|x(t_n)-\al|<\delta.
\]
Next we choose $s_2>t_n$ such that $x(s_2)>\al+\delta$.
We have $t_0\in [s_1,s_2]$ such that
\[
x(t_0)=\min_{s\in [s_1,s_2]}x(s).
\]
Since $t_n\in (s_1,s_2)$ and $x(t_n)<\al+\delta<\min\{x(s_1),x(s_2)\}$, we see that $t_0\in (s_1,s_2)$. This implies
\[
x'(t_0)=0,
\]
which implies $|x'(t_0)|<\delta$. At the same time, we have $x(t_0)-\al\le x(t_n)-\al<\delta$ and $x(t_0)>\al-\delta$ since $t_0>T$. The proof is complete.~\\
\textbf{Case 2:} $\mathcal A(\al)<0$.~\\
Since $\mathcal A$ is continuous, we can find $\delta>0$ such that $\mathcal A(x)<0$ for all $x\in [\al-9\delta,\al+9\delta]$.
By considering the function 
$
z(t)=\al+\delta-x(t)
$ and applying Lemma \ref{son-lem}, we have $T_\delta>0$ such that $x(t)<\al+\delta$ for all $t\ge T_\delta$. Now if there exists $T>T_\delta$ such that $x(t)\ge \al-\delta$ for all $t\ge T$.
In this case, we have $t\in [T,T+2]$ such that $x(T+2)-x(T)=2x'(t)$ and thus $|x'(t)|\le \delta$. If for all $T>T_\delta$, there exists $t>T$ such that $x(t)<\al-\delta$. In this case, we can find $s_1>T_\delta$ such that
$
x(s_1)<\al-\delta
$.
Since $x(t_n)\to \al$ as $n\to\infty$, we can find $t_n>s_1$ such that 
\[
|x(t_n)-\al|<\delta.
\]
Next we choose $s_2>t_n$ such that $x(s_2)<\al-\delta$.
We have $t_0\in [s_1,s_2]$ such that
\[
x(t_0)=\max_{s\in [s_1,s_2]}x(s).
\]
Since $t_n\in (s_1,s_2)$ and $x(t_n)>\al-\delta>\max\{x(s_1),x(s_2)\}$, we see that $t_0\in (s_1,s_2)$. This implies
\[
x'(t_0)=0,
\]
which implies $|x'(t_0)|<\delta$. At the same time, we have $x(t_0)\ge x(t_n)>\al-\delta$ and $x(t_0)<\al+\delta$ since $t_0>T$. The proof is complete.
\end{proof}

\begin{lemma}\label{Trinh-lem}
Let $x:[0,\infty)\to [0,\infty)$ be a continuous function and 
$w:[0,\infty)\to \R$ such that $w(t)\to 0$ as $t\to\infty$. 
Assume that for any $\varepsilon>0$ sufficiently small, there exist 
positive constants $A_\varepsilon, T_\varepsilon$ such that 
\[
x'(t)\le -A_\varepsilon x(t)+w(t)\quad\text{for all } t\ge T_\varepsilon.
\]
Then 
\[
\lim_{t\to\infty} x(t)=0.
\]
\end{lemma}

\begin{proof}
Since $w(t)\to 0$ as $t\to\infty$, we can increase $T_\varepsilon>0$ such that 
\[
|w(t)|\le \varepsilon A_\varepsilon\quad\text{for all } t\ge T_\varepsilon.
\]
This implies that for $t\ge T_\varepsilon$,
\[
x'(t)\le -A_\varepsilon x(t)+\varepsilon A_\varepsilon.
\]
By Gronwall inequality, we have 
\[
x(t)\le x(T_\varepsilon)e^{-A_\varepsilon(t-T_\varepsilon)}
    +\varepsilon\bigl(1-e^{-A_\varepsilon(t-T_\varepsilon)}\bigr).
\]
Hence, if 
\[
t\ge T_\varepsilon+\frac{1}{A_\varepsilon}
          \ln\!\left(\frac{1+x(T_\varepsilon)}{\varepsilon}\right),
\]
we get 
\[
x(t)\le 2\varepsilon.
\]
Since $\varepsilon>0$ is arbitrary small, we conclude $\lim_{t\to\infty}x(t)=0$.
\end{proof}
\begin{lemma}\label{Trinh-lem2}
Let $x:[0,\infty)\to [0,\infty)$ be a continuous function and 
$w:[0,\infty)\to\R$ satisfy $w(t)\to 0$ as $t\to\infty$. 
Assume also that there exists $M>0$ such that 
\[
0\le x(t)\le M \qquad \text{for all } t\ge 0.
\]
Suppose that for every sufficiently small $\varepsilon>0$, there exist 
positive constants $A_\varepsilon$ and $T_\varepsilon$ such that 
\[
x'(t)\ge A_\varepsilon x(t)+w(t)
\qquad \text{for all } t\ge T_\varepsilon.
\]
Then 
\[
\lim_{t\to\infty} x(t)=0.
\]
\end{lemma}

\begin{proof}
Fix any $\varepsilon>0$. By assumption, there exists $T_\varepsilon>0$ such that 
\[
x'(t)\ge A_\varepsilon x(t)+w(t)
\qquad \text{for all } t\ge T_\varepsilon.
\]
Since $w(t)\to 0$ as $t\to\infty$, we may increase $T_\varepsilon$ if necessary so that 
\[
|w(t)|\le \varepsilon A_\varepsilon 
\qquad \text{for all } t\ge T_\varepsilon.
\]
Thus, for all $t\ge T_\varepsilon$,
\[
x'(t)\ge A_\varepsilon x(t)-\varepsilon A_\varepsilon.
\]
Applying Grönwall's inequality, for any $t_1>t_2\ge T_\varepsilon$ we obtain
\[
x(t_1)\ge \bigl[x(t_2)-\varepsilon\bigr] e^{A_\varepsilon(t_1-t_2)}+\varepsilon.
\]
Fix $t_2\ge T_\varepsilon$ and let $t_1\to\infty$.  
Since $x(t)$ is uniformly bounded above on $[0,\infty)$, the right-hand side remains bounded only if
\[
x(t_2)-\varepsilon\le 0.
\]
As this holds for every $t_2\ge T_\varepsilon$, we conclude that $x(t)\to 0$ as $t\to\infty$.
\end{proof}
\begin{lemma}\label{upper}
Consider the triangular dynamical system
\[
\begin{cases}
\partial_t x_1 = F_1(x_1), \\
\partial_t x_2 = F_2(x_1, x_2), \\
\quad\vdots \\
\partial_t x_L = F_L(x_1, x_2, \dots, x_L),
\end{cases}
\]
where each \(F_i: (\R^d)^i \to \R^d\) is continuously differentiable. Let \(x^\star = (x_1^\star, \dots, x_L^\star) \in (\S^{d-1})^L\) be a steady state of this system.
The Jacobian matrix of the linearized system at \(x^\star\) is block upper triangular. Its eigenvalues are determined solely by the diagonal blocks \(D_1 F_1(x^\star), D_2 F_2(x^\star), \dots, D_L F_L(x^\star)\) and are given by the union of their spectra.
\end{lemma}

\begin{proof}
The Jacobian at the steady state \(x^\star\) is given by
\[
J(x^\star)=
\begin{bmatrix}
D_1F_1(x^\star) & 0 & \cdots & 0 \\
D_1F_2(x^\star) & D_2F_2(x^\star) & \cdots & 0 \\
\vdots & \vdots & \ddots & \vdots \\
D_1F_L(x^\star) &  D_2F_L(x^\star) & \cdots & D_LF_L(x^\star)
\end{bmatrix}.
\]
This matrix is block upper triangular. A standard result in linear algebra confirms that the eigenvalues of \(J(x^\star)\) are simply the eigenvalues of its diagonal blocks \(D_iF_i(x^\star)\) for \(1 \le i \le L\).
\end{proof}
\begin{lemma}\label{son-lem-3}
Let $\gamma : [0,\infty) \to \mathbb{R}^d$ be a continuous map, and define the
$\omega$-limit set
\[
K \;=\; \{\, x \in \mathbb{R}^d \;:\; \exists\, t_n \to \infty 
\text{ with } \gamma(t_n) \to x \,\}.
\]
If every point of $K$ is isolated in $K$, then $K$ consists of a single element.
\end{lemma}

\begin{proof}
Fix $x_0 \in K$. We claim that $\gamma(t) \to x_0$ as $t \to \infty$.  
Suppose to the contrary that this convergence fails.  
Then there exist a sequence $s_n \to \infty$ and a number $\delta>0$ such that
\begin{equation}\label{eq:away}
|\gamma(s_n)-x_0| \ge \delta \qquad \text{for all } n .
\end{equation}
Since $x_0 \in K$, for each $n$ we may choose $t_n > s_n$ such that
\begin{equation}\label{eq:close}
|\gamma(t_n)-x_0| < \frac{\delta}{2^n}.
\end{equation}

Fix $n_0 \in \mathbb{N}$.  
By continuity of $\gamma$, for every $n>n_0$ there exists a point 
$w_n \in (s_n, t_n)$ satisfying
\[
|\gamma(w_n)-x_0| = \frac{\delta}{2^{\,n_0}} .
\]
Since the sphere 
$\partial B\!\left(x_0, \frac{\delta}{2^{\,n_0}}\right)$ 
is compact, we may pass to a subsequence (not relabeled) such that 
$\gamma(w_n) \to \alpha$ for some 
\[
\alpha \in \partial B\!\left(x_0, \frac{\delta}{2^{\,n_0}}\right).
\]
Hence
\[
K \cap \partial B\!\left(x_0, \frac{\delta}{2^{\,n_0}}\right) \neq \varnothing .
\]

Since $n_0$ was arbitrary, it follows that $K$ contains points on the spheres
$\partial B(x_0, \delta/2^{\,n_0})$ for infinitely many radii.  
Thus $x_0$ cannot be an isolated point of $K$, contradicting the hypothesis.

Therefore no such sequence $(s_n)$ can exist, and we conclude that 
$\gamma(t)\to x_0$ as $t\to\infty$.  
This shows that $K=\{x_0\}$.
\end{proof}

% \section{Additional Experimental Details}

% \subsection{Ablation Study on Lora weights in~\citep{hu2022lora}}

%%%%%%%%%%%%%%%%%%%%%%%%%%%%%%%%%%%%%%%%%%%%%%%%%%%%%%%%%%%%

% \newpage
% \input{checklist.tex}

\end{document}